\documentclass[11pt,a4paper]{article}
\usepackage[hyperref]{eacl2021}
\usepackage{times}
\usepackage{latexsym}

\usepackage{multicol}
\usepackage{enumitem}
\usepackage{multirow}
\usepackage{svg}
\usepackage{wrapfig}
\usepackage{lscape}
\usepackage{rotating}
\usepackage{epstopdf}
\usepackage{float}
\usepackage{adjustbox}
\restylefloat{table}
\usepackage{booktabs}
\usepackage[capitalize, noabbrev]{cleveref}
\usepackage{xcolor}
\usepackage{microtype}
\newfont{\bitsmaller}{ptmr scaled 950}
\aclfinalcopy % Uncomment this line for the final submission
\def\aclpaperid{789} %  Enter the eacl Paper ID here

\makeatletter
\renewcommand{\paragraph}{%
  \@startsection{paragraph}{4}%
  {\z@}{1.3ex \@plus 1ex \@minus .2ex}{-1em}%
  {\normalfont\normalsize\bfseries}%
}
\makeatother

\title{\textit{Us vs. Them}: A Dataset of Populist Attitudes, News Bias and Emotions}

\author{Pere-Llu\'is Huguet Cabot$^{1,3}$, David Abadi$^2$, \textbf{Agneta Fischer$^2$, Ekaterina Shutova$^1$} \\
 $^1$ Institute for Logic, Language and Computation, University of Amsterdam  \\
  $^2$ Department of Psychology, University of Amsterdam \\
   $^3$ Babelscape Srl, Sapienza University of Rome
  \\
  \texttt{perelluis1993@gmail.com}\\ \texttt{\{d.r.abadi, A.H.Fischer, e.shutova\}@uva.nl}}

\date{}
\begin{document}
\maketitle
\begin{abstract}
Computational modelling of political discourse tasks has become an increasingly important area of research in natural language processing. Populist rhetoric has risen across the political sphere in recent years; however, computational approaches to it have been scarce due to its complex nature. In this paper, we present the new \textit{Us vs. Them} dataset, consisting of 6861 Reddit comments annotated for populist attitudes and the first large-scale computational models of this phenomenon. We investigate the relationship between populist mindsets and social groups, as well as a range of emotions typically associated with these. We set a baseline for two tasks related to populist attitudes and present a set of multi-task learning models that leverage and demonstrate the importance of emotion and group identification as auxiliary tasks.
\end{abstract}

%\color{blue}

\section{Introduction}

Political discourse is essential in shaping public opinion. The tasks related to modelling political rhetoric have thus been gaining interest in the natural language processing (NLP) community. Many of them focused on automatically placing a piece of text on the left-to-right political spectrum. For instance, much research has been devoted to detecting bias in news sources \citep{2019semeval} and predicting the political affiliation of politicians \citep{iyyer-etal-2014-political} and social media users, more generally \citep{conover2011predicting,pennacchiotti2011democrats,preotiuc-pietro-etal-2017-beyond}.
 Other works conducted a more fine-grained analysis, identifying the framing of political issues in news articles \citep{card-etal-2015-media, Yangfeng17}. 
Recently, the field has also turned attention towards modelling the spread of political information in social media, such as detecting fake news or political perspectives \citep{Li2019,ch2020graphbased, Nguyen2020FANGLS}.

Populism has taken the spotlight in political communication in recent years. Various countries around the globe have experienced a surge of populist rhetoric \citep{inglehart2016trump} in both the public and political space. Despite this, approaches to computational modelling of populist discourse have so far been scarce. Due to the flexible nature of populism, annotating populist rhetoric in text is challenging, and the existing research in this area has relied on small-scale analysis by experts \citep{Hawkins2019}. In this paper, we present a new dataset\footnote{Available at \href{https://github.com/LittlePea13/UsVsThem}{https://github.com/LittlePea13/UsVsThem}} of \textit{Reddit} comments annotated for populist attitudes and the first large-scale computational models of this phenomenon. We rely on research in social- and behavioural sciences (e.g., political science and social psychology) to operationalise a definition of populism and an annotation procedure that allows us to capture and generalise the crucial aspects of populist rhetoric at scale. 

In social sciences, populism is essentially described as a not fully developed political ideology and a series of background beliefs and techniques \citep{doi:10.1111/1467-9248.12224}, traditionally centred around the \textit{Us vs. Them} dichotomy. In one of the first attempts to fully define populism \cite{Mudde2004}, it is described as a thin ideology around the distinction between `the people', which includes the `Us', and `the elites' describing the `Them', and with politics being a tool for `the people' to achieve the common good or `the popular will' \citep{Kyle2018,Rodrik19}. Through different platforms, populism uses this rhetoric that revolves around social identity \citep{Hogg2016, abadi2017negotiating} and the \textit{Us vs. Them} argumentation \citep{Mudde2004}.
%Further research has extensively studied populist communication strategies \citep{Rico17} and experimentally demonstrated their effectiveness in inducing emotions \citep{Wirz18}. 
While right-wing populism tends to be characterised by \textit{fear}, \textit{resentment}, \textit{anger} and \textit{hatred}, left-wing populism is associated with \textit{shame} and \textit{guilt} \citep{doi:10.1111/1467-9248.12089,salmela2017}. Moreover, emotions have been shown to be crucial in shaping public opinion more generally \citep{Marcus2002, Marcus2003,demertzis2006emotions, doi:10.1111/1475-6765.12374}.

The design of our annotation scheme and the dataset are inspired by this research, particularly the link between populist rhetoric and both social identity and emotions. Our dataset consists of comments posted on Reddit that explicitly mention a social group. We collect the comments posted in response to news articles across the political spectrum. Through crowd-sourcing, we annotate supportive, critical and discriminatory attitudes towards the group, as well as a range of emotions typically associated with populist attitudes. At the same time, given the relevance of news in the spread of such mindsets, we investigate the relationship between news bias and the \textit{Us vs. Them} rhetoric. Our data analysis reveals interesting interactions between populist attitudes, specific social groups and emotions.

We also present a series of computational models, automatically identifying populist attitudes, based on \textit{RoBERTa} \citep{Roberta19}. We experiment in a multi-task learning framework, jointly modelling \textit{supportive} vs. \textit{discriminatory} attitudes towards a group, the identity of the group and emotions towards the group. We demonstrate that joint modelling of these phenomena leads to significant improvements in detection of populist attitudes.

\section{Related work}
\subsection{Psychology research on populism}

Populist rhetoric revolves around social identity \citep{Hogg2016, abadi2017negotiating,doi:10.1177/1948550617732393,doi:10.1111/1475-6765.12334} and the \textit{Us vs. Them} argumentation \citep{Mudde2004}.
Social identity explores the relations of individuals to social groups. \citet{Turner10} study the evolution of research into social identity and explain the \textit{Us vs. Them} as an inter-group phenomenon, exposing its relation to social identity where the ``self is hierarchically organised and that it is possible to shift from intra-group (`we') to inter-group (`us' versus `them') and vice versa."

Emotions constitute a part of the populist rhetoric and have been essential for information processing and the formation of (public) opinion among citizens \citep{Marcus2002,Nathan05,demertzis2006emotions}.
While social identity and socio-economic factors have been considered primary indicators of populism's growth \citep{Rooduijn18}, emotional factors have lately become a focus within empirical studies, particularly regarding the reactions and spread of populist views \cite{doi:10.1177/0093650216644026}. Specific appraisal patterns have characterised emotions, i.e. an adverse event for which one blames the other is felt as anger - a pattern of appraisals is referred to as \textit{Core Relational Themes} \citep{Smith93, Lazarus2001}, which are the central (therefore core) harm or benefit that underlies each of the negative and positive emotions \citep{Smith93, doi:10.1177/1754073912468165}. Latest attempts to scrutinise populism from the communication science and social psychological perspective have described populist communication and language \citep{doi:10.1177/1748048516640214, Rico17} and demonstrated its operationalisation through experimental research \citep{Wirz18} as being successful in inducing emotions \citep{bakker_schumacher_rooduijn_2021}. 
According to the concept of media populism \citep{Kramer14,Mazzoleni2018}, media effects can further evoke hostility toward the perceived `elites' and (ethnic/religious) minorities, as it contributes to the construction of social identities, such as in-groups and out-groups (i.e., \textit{Us vs. Them}).

\subsection{Modelling political discourse in NLP}

Handcrafted features such as word-frequency \citep{laver_benoit_garry_2003} were initially the base of NLP approaches to model political data.
\citet{DBLP:journals/corr/abs-cs-0607062} introduced the \textit{Convote} dataset of US congressional speeches, and applied an \textit{support-vector machine} (SVM) classifier leveraging discourse information to identify policy stances in it. One of the first uses of neural networks on political text was the work of \citet{iyyer-etal-2014-political}, who used a \textit{recurrent neural network} (RNN) to identify the party affiliation on the \textit{Convote} dataset. 
\citet{Li2019} detected the political perspective of news articles using a \textit{long short-term memory} (LSTM) and a \textit{graph convolutional network} (GCN) on user data from Twitter. Other research investigated the \textit{framing effect} in news articles, which is a mechanism that promotes a particular perspective \citep{entman93}. \citet{card-etal-2015-media} presented the \textit{Media Frame Corpus}, which explores policy framing within news articles. \citet{Yangfeng17} developed a discourse-level Tree-RNN model to identify the framing in each article, by using this corpus dataset. \citet{huguet-cabot-etal-2020-pragmatics} addressed this task by leveraging emotion and metaphor detection in an MTL setup. Other works have also explored sentence-level framing \citep{johnson-etal-2017-leveraging,frame19}.

\textit{Hate speech detection} is not limited to the analysis of political discourses. However, it is related to exposing populist rhetoric in digital communication \citep{Meret17, Estelles20}. Several NLP approaches \citep{mishra2020tackling}, as well as recent shared tasks \citep{zampieri2019semeval, zampieri-etal-2020-semeval} have been proposed to tackle this widespread problem.

While political bias and framing have been widely explored, research on modelling populist rhetoric is still in its nascent stages. Previous work in this area focused on a general description of populism to determine whether a particular text, such as a party manifesto or a political speech contains what is understood as populist rhetoric or attitudes \citep{Hawkins2009,Rooduijn2011,Manucci2017}. Manual annotation was necessary to perform this analysis, often by experts, which also limited the scope and amount of data used, while the resulting datasets are not sufficiently large to train current machine learning models. Furthermore, the description of what constitutes populist rhetoric is still diffuse and covers many different aspects. \citet{Hawkins2019} used holistic grading to assess whether a text is populist or not, to later determine the degree of `populism' of individual political leaders, thus creating the only existing dataset of populist rhetoric, the \textit{Global Populist Database}. 

\section{Dataset creation}
\paragraph{Data collection.}
By annotating Reddit comments that refer to a social group, we monitored how online discussions target them and whether the text showed a positive or negative attitude towards that social group, ranging from support to discrimination. While this process did not ensure capturing the complexity behind the \textit{Us vs. Them} rhetoric, we detected comments directed at certain groups (out-groups) and the attitude towards them within an online community (in-group). We restricted this to six specific groups that populist rhetoric has targeted as an out-group, \textbf{Immigrants}, \textbf{Refugees},
\textbf{Muslims},
\textbf{Jews},
\textbf{Liberals} and
\textbf{Conservatives}. Current research has shown these groups are common targets of populism in the US, UK and across Europe \cite{inglehart2016trump, doi:10.1177/0010414018789490}. Note that to annotate sufficient comments per group we limit the current work to six groups, which is by no means a complete list of targeted groups. We encourage future research to broaden the scope of groups covered.

We chose to extract data from Reddit, (1) due to its availability through the \textit{Pushshift} repository\footnote{\href{https://pushshift.io/}{https://pushshift.io/}} \citep{baumgartner2020pushshift} and the \textit{Google Bigquery} service, (2) its social identity dynamics (in-group vs. out-group) as close-nit communities created by sub-Reddits, (3) its nature as a social news aggregation platform, and (4) that it has been shown to encourage toxic communication between users and hate speech towards social groups \cite{doi:10.1177/1461444815608807, 10.1371/journal.pone.0228723, Munn2020}.
 To filter the data for annotation, we followed several steps. (1) We identified submissions in Reddit which shared a news article from a news source listed at the \textit{AllSides} website\footnote{\href{https://www.allsides.com/unbiased-balanced-news}{https://www.allsides.com/unbiased-balanced-news}}, (2) we extracted comments which are direct replies to the submission where both the news article title and the comment match any of the keywords for our groups. Keywords were devised using online resources from the  \textit{Anti Defamation League}\footnote{\href{https://www.adl.org/}{https://www.adl.org/}} as well as by consulting social scientists. The full list of keywords can be found in Appendix \ref{sec:appendix} Table \ref{tab:keywords}. (3) We selected comments with a minimum of 30 words and a maximum of 250 words, and sampled from specific periods during which each group was actively discussed on Reddit. See Appendix \ref{sec:appendix} Table \ref{tab:times} for details. (4) We removed comments that contained keywords from multiple social groups to make the annotation process more straightforward. 
(5) We randomly sampled 300 comments per group and news source bias according to \textit{AllSides} (left, centre-left, centre, centre-right, right), resulting in a total of 9000 Reddit comments. Note that the bias is not directly related to individual comments, but rather to the news article the comment responded to. 

\paragraph{Annotation procedure.}

To capture the \textit{Us vs. Them} rhetoric, we asked: \textbf{What kind of language does this comment contain towards \textit{\_group\_}?}\label{Mturk1},
where \textit{\_group\_} corresponds to the specific social group that comment refers to. Respondents had four options: \textbf{Discriminatory}, \textbf{Critical}, \textbf{Neutral} or \textbf{Supportive}. An extended description and an example as presented to annotators can be found in Appendix \ref{sec:mturkq1}, and Figure \ref{fig:questionusvsthem}. We asked annotators a second question to capture the emotions expressed towards the group in the same comment. We extended Ekman's model of 6 \textit{Basic Emotions} \citep{Ekman1992} to a 12-emotions model, which includes a balanced set of positive and negative sentiments. Specifically, we included emotions previously shown to be associated with populist attitudes \citep{demertzis2006emotions}. We also provided the annotators with a brief description of each emotion, inspired by the concept of \textit{Core Relational Themes} \citep{Lazarus1990}.
The positive emotions are \textbf{Gratitude}, \textbf{Happiness}, \textbf{Hope}, \textbf{Pride}, \textbf{Relief} and \textbf{Sympathy}, and the negative emotions are \textbf{Anger},
\textbf{Fear},
\textbf{Contempt},
\textbf{Sadness},
\textbf{Disgust} and
\textbf{Guilt}. Detailed descriptions can be found in Appendix \ref{sec:mturkq1} along with an example \ref{fig:questionemotions}. The annotation was conducted on \textit{Amazon Mechanical Turk }(MTurk) and its framework can be accessed \href{https://littlepea13.github.io/mturk_annotation/}{here}. 

\paragraph{Annotation reliability.}

Once the MTurk annotation was completed, we deployed the \textit{CrowdTruth 2.0} toolkit \citep{dumitrache2018CrowdTruth} to assess the quality of annotations and to identify unreliable workers. \textit{CrowdTruth} includes a set of metrics to analyse and obtain probabilistic scores from crowd-sourced annotations. 
\textit{Worker Quality Score} (WQS) is a metric that measures each worker's performance, by leveraging their agreement with other annotators and the difficulty of annotation. The \textit{Media Unit Annotation Score} (UAS) is given for each comment and possible answer, indicating the probability with which each option could be the gold label. Finally, \textit{Media Unit Quality score} (UQS) describes the quality of annotation for each comment. 
We removed annotations by workers with a WQS lower than $0.1$ and those with high disagreement after manually checking their responses, and recomputed the metrics. We removed comments left with only one annotator and comments with a UQS lower than $0.2$. This resulted in 4278 comments with 5+ annotators, and 2564 comments with less, which constitute our final dataset.

Following the same procedure as \citet{demszky2020goemotions}, we computed inter-rater correlation \citep{Tibau19} by using Spearman correlation. We took the average of the correlation between each annotator's answers and all other annotators' average answers that labelled the same items. We obtain a range between 0.5 and 0.13 per emotion. The lowest agreement is for \textit{Relief} (0.13), in line with \citet{demszky2020goemotions}, where the lowest value for correlation agreement being 0.16, and 0.17 for \textit{Relief}. The full distribution can bee found in \cref{app:reliability} \cref{fig:correla_dist}.

\begin{figure}[!t]
    \centering
    \adjustbox{width=\columnwidth,set height=3.7cm,set depth=3.5cm,frame,center}{
\resizebox{0.9\textwidth}{!}{\begin{tabular}{llll}

\multicolumn{4}{p{0.53\textwidth}}{\textit{Of course Dems are stealing the elections. They are playing by a different set of rules - being ruthless and violent. Dems take no prisoners and show no mercy to their enemies. The sooner GOP realizes it, the better. Because we have to up our game. If things go this way, we have only one way to save this country - Martial Law and kick every single liberal out!}}  \\
%\midrule
 \textbf{Label} &\textbf{UsVsThem} & \textbf{Group} & \textbf{Emotions}% & \textbf{News source} & \textbf{Source bias}
 \\ 
%\midrule
 Discriminatory &$1$ & Liberals & Contempt, %& The Federalist & right 
\\
& & \multicolumn{2}{r}{Disgust \& Fear}\\
\midrule
\multicolumn{4}{p{0.53\textwidth}}{\textit{You do realize it's sad to celebrate the US cutting the number of refugees down, right? These are people who come here seeking a safe haven from the violence or despair of their home countries, and we're turning them away. 
}}  \\
%\midrule
  \textbf{Label} &\textbf{UsVsThem} & \textbf{Group} & \textbf{Emotions} %& \textbf{News source} & \textbf{Source bias}
 \\ 
%\midrule
 Supportive & $0$  & Refugees & Sympathy %& Washington Times & centre-right 
\\
\end{tabular}}
    }
    \caption{Two samples of our \textit{Us Vs. Them dataset.}}
    \label{fig:dummy}
\end{figure}
\section{Data analysis}
\label{sec:data_analysis}
\subsection{UsVsThem scale}
For the \textit{Us vs. Them} question, we aggregated the answers into a continuous scale. To obtain a score for each comment, we computed the \textit{CrowdTruth} UAS for each of the labels assigned to it and then take a weighted sum. We assigned the weight of $0$ to the \textit{Supportive} label, $1/3$ to \textit{Neutral}, $2/3$ to \textit{Critical} and $1$ to \textit{Discriminatory}. %Using \textit{CrowdTruth} UAS, we average each dimension by those values to get the score of each comment in the continuous scale.
The frequency distribution of comments on this scale can be seen in \cref{sec:appanalysis} \cref{fig:UsVsThemRegression}. From now on, we will refer to it as the \textit{UsVsThem} scale.

The scale is skewed, with an overall mean of $0.551\pm 0.265$. Although our data selection was random across the selected news sources and groups, there are more comments with \textit{negative} attitudes towards selected groups than positive or neutral ones due to its nature and our keyword selection.

We performed a two-way ANOVA (\textit{Analysis of Variance}) test \citep{FUJIKOSHI1993315} on news bias and social groups as independent variables and the \textit{UsVsThem} scale as the dependent variable to see whether the interactions between groups and news bias are significant. 
One-way tests show statistical significance.
Interestingly, there was a statistically significant interaction between the effects of social groups and bias on the \textit{UsVSThem} scale, $F (1, 20) = 12.33, p < 0.05$. Values can be found in \cref{sec:appanalysis} \cref{tab:Anova2way}. Therefore, we explored the interaction between them and the influence of news bias on how each group is perceived. We performed a \textit{Tukey HSD test} to check for significance between means in the \textit{UsVsThem} scale.

\begin{figure}[t!]
    \centering
    \def\svgwidth{\columnwidth}
    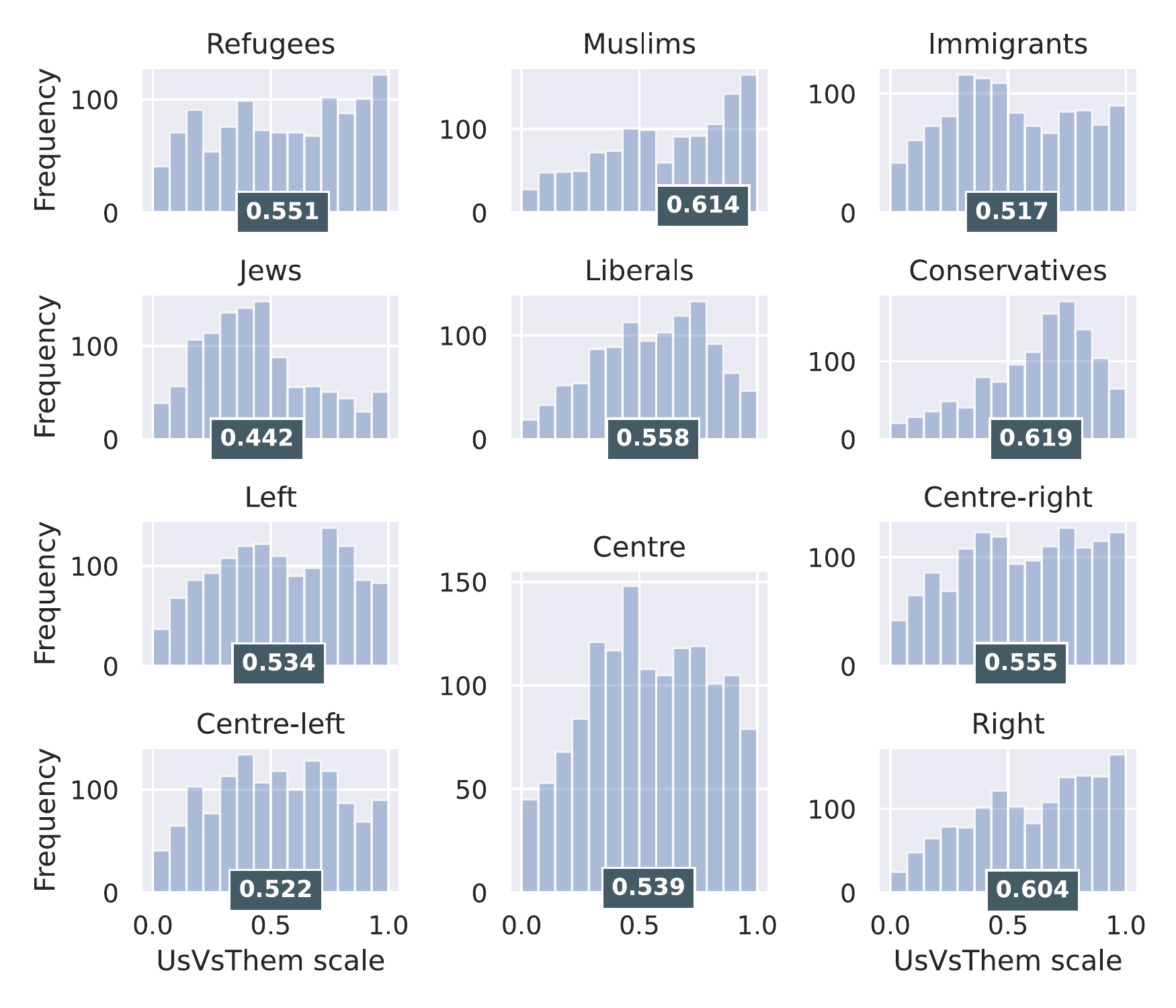
    \caption{Comment frequency distribution on the \textit{UsVsThem} scale per social group and news source bias. The mean for the scale is shown at the x axis.}
    \label{fig:UsVsThemRegressionGroupsBias}
\end{figure}

\paragraph{Social groups.}
There were differences between groups in terms of the \textit{UsVsThem} scale when looking at the comment frequency distributions in Figure \ref{fig:UsVsThemRegressionGroupsBias}. 
For \textit{Refugees}, the distribution was relatively flat as they received a similar amount of positive and negative attitude comments. 
\textit{Immigrants} showed a similar distribution with fewer comments in the higher end, i.e. the group received less discrimination than \textit{Refugees}. Despite the two share many inherent similarities, these differences may be explained by negative media coverage of \textit{Refugees} portrayed as a threat and being attributed to negative attitudes. \textit{Muslims} 
received a higher amount of discriminatory comments than any other group. On the other hand, \textit{Conservatives} showed a similar mean, due to a very high amount of critical comments. \textit{Liberals} also received a relatively high amount of critical comments. 
Both share moderately low tails, as they received less support and discrimination. 
Finally, \textit{Jews} showed lower critical and discrimination values, with most values around \textit{Neutral},
having the lowest mean value of all social groups.
These variations translate into a significant ($p<0.05$) difference between the means of each group, except for \textit{Conservatives} and \textit{Muslims}, %($p=0.74$) 
and for \textit{Liberals} and \textit{Refugees}. 

\paragraph{News source bias.}

In this case, the bias was not directly associated with the comment itself. However, differences in the distribution of comments and the out-group attitudes based on the original article's bias can be observed, as shown in Figure \ref{fig:UsVsThemRegressionGroupsBias}. Moreover, means increased from the \textit{centre-left} to the \textit{right} bias. 
Interestingly, there was no symmetry at the \textit{centre} bias, contrary to the \textit{Horseshoe Theory} \citep{doi:10.1177/1948550618803348}, which argues both ends of the political spectrum closely resemble one another. 
In terms of significant differences, all biases were significantly different from the \textit{right} bias ($p<0.05$), and there was a significant difference between \textit{centre-left} and \textit{centre-right} ($p<0.05$). The remaining groups showed no significant difference.

\paragraph{Groups and news source bias.}

In line with the above-mentioned bias effect, there was almost always a significant difference between \textit{right} and \textit{centre-right} bias and the rest for each group. 
Only \textit{right} bias showed a distinct high value and a negative attitude towards \textit{Immigrants}, which even exceeds those towards \textit{Refugees}. 
With the exception of the attitude towards \textit{Conservatives}, \textit{centre}, \textit{centre-left} and \textit{left} showed lower degrees of negative attitude towards any of the groups. 
Full results can be seen in \cref{sec:appanalysis} \cref{tab:biasRegression,fig:UsVsThemRegressionGroupsBiasCombined}.

\subsection{Emotions}
Instead of using \textit{CrowdTruth} for emotions, we considered an emotion as being present in the comment provided that at least 1/4 of annotators selected it. In case more than half of annotators marked that comment as \textit{Neutral}, it was labelled as \textit{Neutral}. This way, a comment can contain more than one emotion, except for \textit{Neutral}. Unless specified otherwise, in this subsection \textit{Neutral} refers to emotionally neutral.

\begin{figure}[t!]
    \centering
    \includegraphics[width=1\columnwidth]{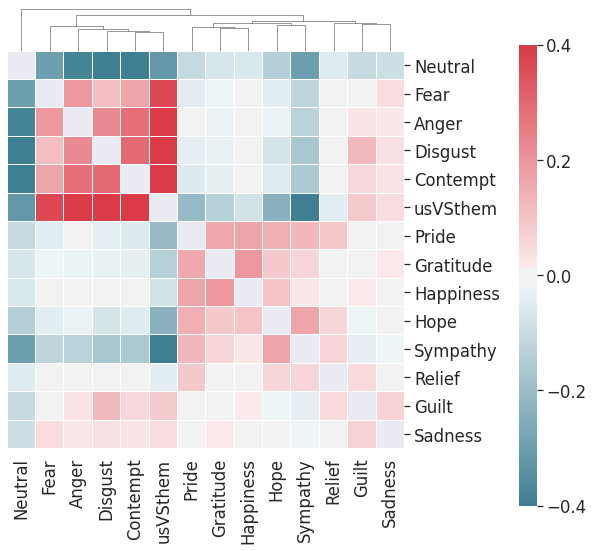}
    \caption{Correlation heat-map for different emotions.}
    \label{fig:emotionsCorr}
\end{figure}
In Figure \ref{fig:emotionsCorr} we present the correlations between the values for each emotion dimension and the \textit{UsVsThem} scale across all comments, in the same fashion as \citet{demszky2020goemotions}. We show the hierarchical relations at the top, demonstrating which emotions interact more strongly with each other. 
The frequency of emotions in our dataset is as follows:
\textit{Anger}               1724,
\textit{Contempt}            2538,
\textit{Disgust}             1843,
\textit{Fear}                1136,
\textit{Gratitude}             70,
\textit{Guilt}                170,
\textit{Happiness}             59,
\textit{Hope}                 307,
\textit{Pride}                174,
\textit{Relief}                37,
\textit{Sadness}              122,
\textit{Sympathy}            1139,
\textit{Neutral}             2094.

\paragraph{Emotions and the \textit{UsVsThem} scale.}

We were interested in the interaction between emotions and social identity by exploring how the \textit{UsVsThem} scale is shaped for each emotion. Not surprisingly, comments with negative emotions showed a higher value on the \textit{UsVsThem} scale and a high correlation with it, except for \textit{Guilt} and \textit{Sadness}. \textit{Contempt} showed the strongest correlation, while \textit{Anger} and \textit{Fear} showed a higher proportion of \textit{Discriminatory} comments. \textit{Guilt} and \textit{Sadness}, on the other hand, are characterised by a lower amount of comments in the \textit{Discriminatory} range.
In line with these results, the \textit{UsVsThem} scale had a negative correlation with \textit{Sympathy} and \textit{Neutral} comments, and while \textit{Sympathy} was the most frequent positive emotion, other emotions displayed a very similar relation to the \textit{UsVsThem} scale. These results are visualised in Figure \ref{fig:emotionsCorr} including the \textit{UsVsThem} scale and the distributions are summarised in Figure \ref{fig:emotionsUsVsThem} in \cref{sec:appanalysis}.

\paragraph{Emotions and groups.}

We used a two-sided proportion z-test to check for significant differences since emotions are discrete variables. More than 25\% of comments towards \textit{Muslims} and \textit{Refugees} showed \textit{Fear}, with no significant difference between the two, followed by \textit{Immigrants} at 21.5\% and other groups at less than 10\%. Another notable finding is that \textit{Contempt} (47.7\%) and \textit{Disgust} (44.4\%) were significantly higher towards \textit{Conservatives}, particularly the latter, which for other groups never exceeded 30\% of comments. \textit{Sympathy} for \textit{Liberals} (6.5\%) and \textit{Conservatives} (6.9\%) was significantly lower when compared to other groups. 
\textit{Hope} was present in a significantly higher number of comments for \textit{Liberals} (7.9\%). 
The values for all proportions can be found in Appendix \cref{sec:appanalysis} \cref{tab:groupsemotions}.

\paragraph{Emotions and bias.}
Not many differences between biases were found. Most salient was the \textit{right} bias showing a higher value in all negative emotions, significantly for \textit{Anger} (31.5\%), \textit{Contempt} (43.4\%) and \textit{Fear} (21.9\%). All proportion values can be found in \cref{sec:appanalysis} \cref{tab:biasemotions}.

\section{Modelling populist rhetoric}
\subsection{Main tasks}
Our models' main focus was to assess to which degree a social group is viewed as an out-group and whether in a negative or discriminatory manner. Our annotation procedure provided a scale from \textit{Supportive} to \textit{Discriminatory} for each comment. While this scale is artificial and highly dependent on our task's context, it provides a good indication of how strongly a social group is targeted in social media comments.

\paragraph{Regression \textit{UsVsThem}.} In our models, we explored two different main tasks. The first task was to predict the values on the \textit{UsVsThem} scale in a regression model. This scale provides a score for each comment, which illustrates the attitude towards a social group mentioned in the comment ranging from \textit{Supportive} (closer to 0) to \textit{Discriminatory} (closer to 1). Values in between depict an intermediate attitude, \textit{Neutral} lies at $1/3$, and \textit{Critical} at $2/3$. By predicting the score, we modelled the out-group attitude of each comment. We used $33\%$ of the data as the test set, and $13.4\%$ as the validation set.

\paragraph{Classification \textit{UsVsThem}.} Our second task was to classify each comment in a binary fashion as whether the comment shows a negative attitude towards a group, i.e., \textit{Critical} or \textit{Discriminatory}, or not, i.e., \textit{Neutral} or \textit{Supportive}. This task resulted in a relatively balanced dataset, with $56\%$ of \textit{Critical} or \textit{Discriminatory} comments. We used the same splits as before.

\subsection{Auxiliary tasks}
\paragraph{Emotion detection.} 
Interactions between populist rhetoric and emotions have been explored in political psychology through surveys and behavioural experiments \citep{Fischer2007BeatTO, Tausch11,doi:10.1177/0539018417734419,doi:10.1080/17457289.2018.1441848,ROLLWAGE2019820,nguyen_2019,doi:10.1177/1532673X19857968}. %, AbadiSurvey}. 
 This is consistent with our findings in section \ref{sec:data_analysis} and further motivates modelling emotions in the context of populist rhetoric.
For each comment, emotions were annotated as a Boolean vector. For our task, some emotions were rarely annotated or only present alongside more frequent ones. They increased the difficulty of the task while not providing relevant information. To simplify the auxiliary task, we considered the 8 most common emotions, \textit{Anger, Contempt, Disgust, Fear, Hope, Pride, Sympathy} and \textit{Neutral}.

\paragraph{Group identification.} 
In the work of \citet{burnap2016us}, types of hate speech were differentiated based on race, religion, etc., and models were trained specifically on those categories. In \citet{elsherief2018hate}, data-driven analysis of online hate speech explored in profundity the differences between \textit{directed} and \textit{generalised} hate speech, and \citet{ICWSM1613147} analysed the different targets of hate online.
In our case, the \textit{Us vs. Them} rhetoric metric showed significant differences for each group as we have seen in the previous section. Therefore, we hypothesised that the information bias \citep{Caruana93multitasklearning} the group identification task provides will help understand the \textit{Us vs. Them} rhetoric aimed at the different social groups, which motivated its role as an auxiliary task.

\subsection{Model architecture}

We used the \textit{Robustly Optimized BERT Pretraining Approach} (RoBERTa) \citep{Roberta19} in its \textsc{base} variant as provided by \citet{wolf2019huggingfaces}. 

\paragraph{Multi-task learning.}
In all setups, tasks shared the first eleven \textit{transformer} layers of \textit{RoBERTa}. The final 12th layer was task-specific, followed by a classification layer that used the hidden representation of the \texttt{<s>} token, to output a prediction.
We used scheduled learning, where the losses of each task are weighted and changed during training. We also experimented with a three-task MTL model where the two auxiliary tasks are learned simultaneously.

We assigned three different loss weights associated with each task, $\lambda_{m}$ for the main task, either regression or binary classification; $\lambda_{e}$ for emotion detection; $\lambda_{g}$ for group identification. For MTL with one auxiliary task, $\lambda_{m} + \lambda_e = \lambda_m +  \lambda_g = 2$, while for the three-task MTL: $\lambda_m + \lambda_e +\lambda_g = 3$.
 
\paragraph{Regression \textit{UsVsThem}.} 
We used \textit{Mean Squared Error} loss with a sigmoid activation function for the main task. For emotion identification as the auxiliary task, we used \textit{Binary Cross-Entropy} loss, and for the group identification, we used \textit{Cross-Entropy} loss, both with sigmoid activation. For all MTL models, there was a warm-up period of $\omega$ epochs, after which the weight is changed to $\lambda_g = 10^{-2}$ and $\lambda_e = 10^{-5}$, and $\lambda_e = \lambda_g = 10^{-5}$ for the three-task setting.

\paragraph{Classification \textit{UsVsThem}.} 
We used \textit{Cross-Entropy} loss with a sigmoid activation function for the main task. The remaining tasks were kept the same as with the Regression case above. For all MTL models, there was a warm-up period of $\omega$ epochs, after which the weight was changed to $\lambda_g = 10^{-2}$ and $\lambda_e = 10^{-2}$, and $\lambda_e = \lambda_g = 10^{-5}$ for the three-task setting.

\subsection{Experimental setup}
\paragraph{Regression \textit{UsVsThem}.} 
We report model performance in terms of Pearson correlation coefficient (R). We found the optimal STL hyperparameters using the validation set: a learning rate of $3e-05$, a lineal warm-up period of 2 epochs and dropout of $0.15$. The batch size used was 128. 
These hyperparameters were kept constant across our experiments for the regression \textit{UsVsThem} task. 
For the emotion detection MTL setup, 
 $\lambda_e = 0.15$ and $\omega = 8$.
For the groups MTL, $\lambda_g = 0.15$ and $\omega = 5$.
For the three-task MTL model we obtained optimal validation performance by setting $\omega = 8$ and both $\lambda_g = \lambda_e = 0.073$, which was the equivalent of $\lambda_g = \lambda_e = 0.05$ for the two-task MTL.

\paragraph{Classification \textit{UsVsThem}.} Similarly, we ran a grid-search to find the best hyperparameters for the classification setup. For the STL model, 
we obtained a learning rate of $5e-05$, a warm-up of 2 epochs and an extra dropout of $0.2$.
For emotions-MTL, 
 $\lambda_e = 0.2$ and $\omega = 8$.
For the groups-related MTL,  
 $\lambda_g = 0.25$ and $\omega = 5$.
For the three-task MTL, $\lambda_e = 0.95$, $\lambda_g = 0.25$, and $\omega = 8$. For both regression and classification, we report performance averaged over 10 different seeds.
\subsection{Results}
\begin{table*}[t!] \centering
\resizebox{0.78\textwidth}{!}{
\begin{tabular}{lcccc}
\toprule
& \textbf{STL} & \textbf{MTL, Emotion} & \textbf{MTL, Group} & \textbf{MTL, Emotion \& Group}\\ 
\midrule
Pearson R                                     &  0.545 $\pm$ 0.005        & \textbf{0.553 $\pm$ 0.009}          &   \textbf{0.557 $\pm$ 0.012}     &  \underline{\textbf{0.570 $\pm$ 0.009}} \\ 
Accuracy                           &0.705 $\pm$ 0.006 &0.710 $\pm$ 0.009 &0.711 $\pm$ 0.007 &  \textbf{0.717 $\pm$ 0.004} \\
\bottomrule
\end{tabular}}
\caption{Results for the \textit{Us vs. Them} rhetoric as regression and classification tasks. Significance compared to STL is bolded ($p<0.05$). Significance compared to two-task MTL is underlined ($p<0.05$). Average over 10 seeds.}
\label{tab:resultsUsVsThem}
\end{table*}
Results are presented in Table \ref{tab:resultsUsVsThem}. We find that MTL outperforms STL in both versions of our task. 
\paragraph{Regression \textit{UsVsThem}.} The STL baseline showed a 0.545 Pearson R to the gold score. When emotion identification was used as an auxiliary task, the performance increased by almost one point, to 0.553. The groups MTL setup showed a higher increase, up to 0.557. Both improvements were significant compared to the STL model, using the Williams test \citep{williamstest}. Perhaps all the more interesting is that the three-task MTL model achieved the highest performance, even without its hyperparameters being specifically tuned as with the other setups. It resulted in a Pearson R of 0.570, i.e. over 2 points performance increase over STL, a statistically significant improvement over both STL and the remaining two MTL approaches.
\paragraph{Classification  \textit{UsVsThem}.} Although not shown in the table, the accuracy baseline for a majority class classifier would be 0.550. All models highly surpassed that, with the real baseline set by the STL setup achieving a 0.705 accuracy. Results for the MTL approaches were similar to what we observed in the regression task. Emotion-MTL increased performance by half a point, to 0.710, as did group-MTL, with 0.711. %Both improvements are significant according to the permutation test.
The best performing model was again the three-task MTL, at 0.717, yielding a statistically significant improvement over STL, using the permutation test.

\subsection{Analysis}
\begin{table*}[!h]
% \bitsmaller 
\centering
% Through the resizebox you ensure that the table is the same width as the column, to avoid going slightly outside the margins triggering a desk reject
\resizebox{1\textwidth}{!}{\begin{tabular}{llllllll}
\toprule
$n$ &\multicolumn{7}{l}{\textbf{Reddit Comment}} \\ 
\midrule
1 & \multicolumn{7}{p{1.1\textwidth}}{Does anyone else think it's absurd that we have Muslims serving in our congress? Literal foreign agents acting as if they hold the interests of the country and its people at heart? They never talk about the will of the people. It's always some bullshit about how white men (who founded and built this country) are evil and we need to let in more 3rd worlders who want to bomb and kill us. This is literal clown world tier nonsense.}\\
\midrule
\midrule
2 & \multicolumn{7}{p{1.1\textwidth}}{You proud of yourselves, making 3 year olds represent themselves in immigration court? You fucking proud of that insanity? All for the sake of keeping out a gang that has already been in America for a long time, meanwhile regular home grown white kids are murdering dozens of their own classmates but goddam, at least they we're legal, amirite}  \\
\midrule
\midrule
3 & \multicolumn{7}{p{1.1\textwidth}}{Conservatives have every right to revolt. If we don't get our way we will destroy the country. I hope the left keeps pushing us to provoke a civil war. Or maybe Commiefornia should secede. Maybe that's the best thing that can happen, a complete break up. That way we can have our ethnostate, and the left can have their degenerate cesspool without us paying taxes for it. The US is dead anyway. It's time to burn this diverse shithole to the ground. It will be the ultimate proof that diversity doesn't work.}\\
\midrule
 &\textbf{Label}& \textbf{MTL, E. \& G.} & \textbf{MTL, Emo.} & \textbf{MTL, Groups} &	\textbf{STL} & \textbf{Group} & \textbf{Emotions}\\ 
\midrule
1 & $1.000$ & $0.872$ & $0.870$ & $0.847$ & $0.759$ & Muslims & Anger, Contempt, Disgust \& Fear\\
% &  &  &  &  &  &  & Disgust \& Fear \\
\midrule
2 & $0.02$ & $0.774$ & $0.874$ & $0.740$ & $0.834$ & Immigrants & Sympathy \\
\midrule
3 & $0.071$ & $0.729$ & $0.773$ & $0.747$ & $0.8$ & Conservatives & Hope \& Pride \\
\bottomrule
\end{tabular}}
\caption{Examples of predictions for comments. Predictions are averages over 10 seeds for each model.}
\label{tab:examplesBody}
\end{table*}
%Here we focused on the regression task as it expresses the task with more complexity on each comment. 
\paragraph{Qualitative and error analysis.} We selected comments with higher values on the scale where MTL improved the STL baseline predictions for the regression task. Comments with high emotion valence were better predicted by models that included emotion identification. Comments that had group-specific rhetoric with references to (derogatory) terms such as `illegal aliens' were better predicted by models that incorporated group identification (see the first example in Table \ref{tab:examplesBody}).

The standard deviation of the difference between the STL and the three-task predictions was just $0.055$. This means that MTL helped capture nuanced information that improved prediction; however, comments with high squared error for STL still showed similar behaviour for MTL models. This aspect is shown in Appendix \ref{sec:appendixresults} Figure \ref{fig:errors_stl_mtl}. % as a correlation between the squared error measures of the two models. 
All models' squared error showed a pair-wise Pearson correlation higher than $0.92$. 
This observation prompted us to investigate comments with a high squared error. We identified three different sources. (1) \textbf{Comments with emotionally charged language}, slurs, or insults, which may often be associated with a more negative attitude towards a group, were mispredicted due to not being negative towards such group or being used ironically or satirically. 
(see second example in Table \ref{tab:examplesBody}). 
(2) \textbf{Reference to multiple groups}: we removed comments that included keywords from similar groups, however it was impossible to account for all the terms that may refer to other groups. %or groups that weren't in our annotation procedure. 
Hence, there are comments for which the prediction seems to be about a target different than the one at annotation time (see the third example in Table \ref{tab:examplesBody}). More examples can be found in \cref{sec:appendixresults} \cref{tab:examplesTotal}.
(3) \textbf{Annotation error} is expected in any crowd-sourced annotation. 
While these were not as frequent as to pose a problem during training, they did occur as incorrect model predictions that can mistakenly decrease performance.

\paragraph{Analysis of model representations.}
\begin{figure}[t!]
    \centering
    \includegraphics[width=\columnwidth]{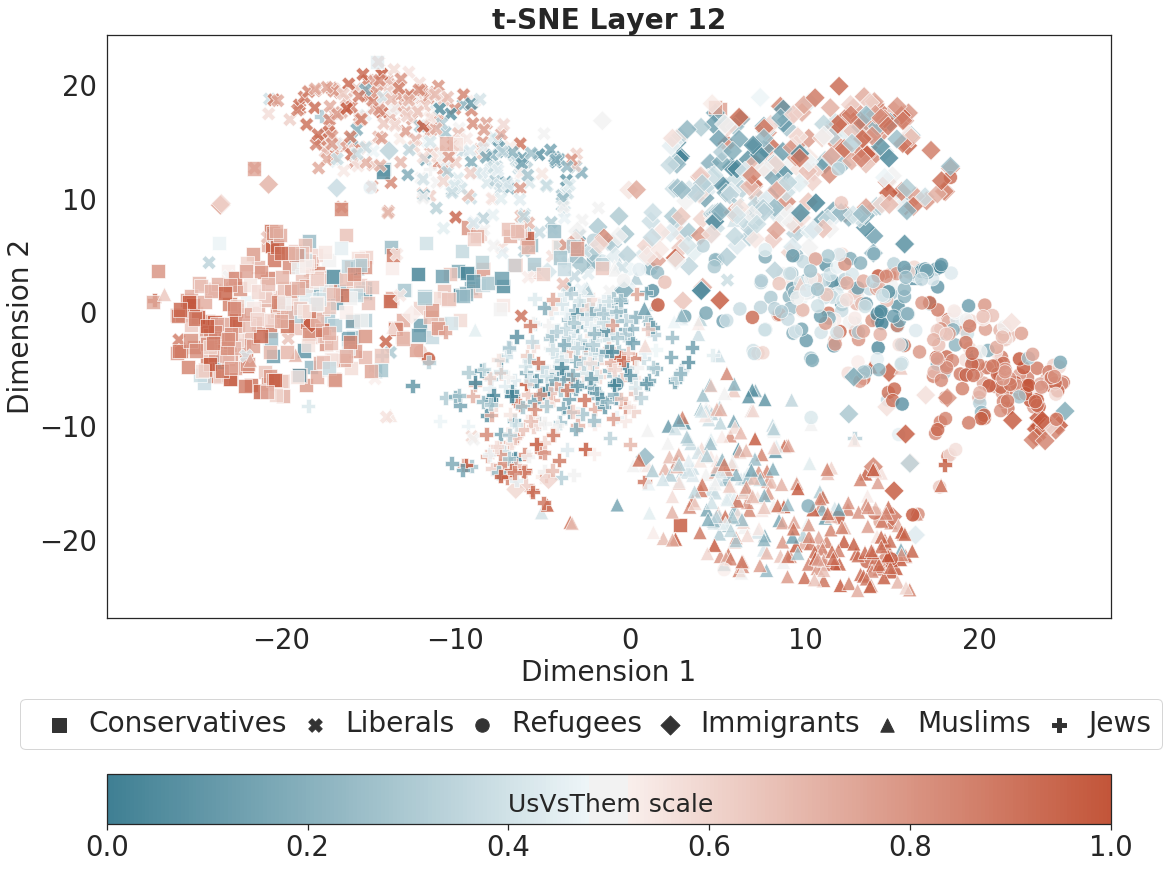}
    \caption{Three-task MTL main task specific layer.}
    \label{fig:small_mtl_three}
\end{figure}

Our qualitative analysis showed that auxiliary tasks had a positive effect on predictions. Still, it cannot explain how the model changes its underlying structure and Reddit comments' encoding. 
We explored how auxiliary tasks affected how the network encodes information through its layers to get a better understanding.
We %did not 
used %any classifiers except for 
\textit{t-Distributed Stochastic Neighbour Embedding} (t-SNE) \citep{maaten2008visualizing}, a stochastic technique for dimensionality reduction focused on high dimensional data visualisation. We used it 
to visualise the hidden representations of the test set comments in-between \textit{transformer} layers across the network. 
We present the results for both STL and the three-task MTL in \cref{sec:appendixresults} \cref{fig:layers_stl,fig:layers_three}, where for both the first layers showed some structure not related to the tasks at hand. As we were using pre-trained weights from \textit{RoBERTa}, this could be explained by the first layers modelling lower-level language characteristics as shown empirically in \citet{Tenney19}, where probing mechanisms indicate early layers being more relevant for tasks such as \textit{POS tagging}. 
For STL, the last layers showed the \textit{UsVsThem} scale continuously in the y axis. 
Once we introduced the auxiliary tasks of group identification and emotion classification differences in the last layers were exacerbated. 
For three-task MTL the last layers showed clusters for each social group, and related groups were closer together, such as \textit{Refugees} and \textit{Immigrants}, or \textit{Liberals} and \textit{Conservatives}. We also observe a radial distribution with highly emotional comments further away from the centre. 
Comments with very distant values on the scale (\textit{Discriminatory} and \textit{Supportive}) were closer together than with those in the mid-range (\textit{Neutral} and \textit{Critical}) as seen in \cref{fig:small_mtl_three} and \cref{sec:appendixresults} \cref{fig:final_layer_main}. While paradoxical, our interpretation is that the model leverages the valence of emotion, where \textit{Discriminatory} and \textit{Supportive} comments are more loaded with emotion. This leads to a better performance of MTL compared to STL. This idea is supported by the distribution of emotions on the last layer, where emotionally neutral comments are closer to the centre of the plot, while more emotionally charged comments radially increase, visualised in \cref{sec:appendixresults} \cref{fig:final_main_emo}. In \cref{sec:appendixresults} \cref{fig:final_group_emo,fig:layers_emotions} we present the emotion distribution for the group and emotion-specific layers, respectively.

\section{Conclusions}
We presented a new, large-scale dataset of populist rhetoric and the first series of computational models on this phenomenon. We have shown that joint modelling of emotion and populist attitudes towards social groups enhances performance over the single-task model, further corroborating previous research findings in various social sciences. Future work may deploy social information (e.g., Twitter) or explore the interactions of populist attitudes and the political bias of news articles as provided in our \textit{Us Vs. Them} dataset.
%\section*{Acknowledgments}

\paragraph{Funding statement.} This research was funded by the European Union’s H2020 project \textit{Democratic Efficacy and the Varieties of Populism in Europe} (DEMOS) under H2020-EU.3.6.1.1. and H2020-EU.3.6.1.2. (grant agreement ID: \href{https://cordis.europa.eu/project/id/822590}{822590}) and supported by the European Union’s H2020
Marie Sk\l{}odowska-Curie project \textit{Knowledge Graphs at Scale} (KnowGraphs) under H2020-EU.1.3.1. (grant agreement ID:
\href{https://cordis.europa.eu/project/id/860801}{860801}).
\bibliography{anthology,eacl2021}
\bibliographystyle{acl_natbib}
\clearpage
\appendix

\section{Supplemental material}
\label{sec:appendix}
\subsection{Data collection}
\label{sec:annotation}

\begin{table}[h!]
\centering
\resizebox{\columnwidth}{!}{\begin{tabular}{lll}\toprule
              & Time ranges                                                                                                                           & Events                                                                                   \\ \midrule
Conservatives & 2016/09/15 - 2016/12/15& Election periods \\ & 2018/09/15 - 2018/12/15   \\
Liberals      & 2016/09/15 - 2016/12/15& Election periods\\ & 2018/09/15 - 2018/12/15      \\
Muslims       &  2016/11/01 - 2017/11/30 & Trump Muslim ban,\\ & 2018/04/01 - 2018/05/01 & Mosque attacks.\\ & 2019/03/01 - 2019/06/01                \\
Immigrants    &  2016/11/01 - 2017/11/30 & Migrant caravans,\\ & 2017/01/15 - 2017/03/15 & Children at the US\\ & 2018/06/17 - 2018/07/01 & border\\ & 2018/10/01 - 2019/02/01  \\ Jews & 2018/10/20 - 2018/11/25 & Christchurch shooting\\ \midrule
\end{tabular}}
\caption{Events and periods used for each group. If comments were not sufficient, they were sampled randomly from other time ranges. Refugees did not have enough overall comments to be filtered by time range.}
\label{tab:times}
\end{table}
\begin{table}[ht!]
\centering
\resizebox{\columnwidth}{!}{\begin{tabular}{lll} \toprule
              & News Title                                                              & Comment                                                                \\ \midrule 
Refugees      & refugee, asylum seeker                                                  & refugee, asylum seeker, \\& & undocumented, colonization                      \\
Immigration   & -migra-, undocumented,                                     & -migra-, undocumented,                                     \\
   & colonization                                     & colonization                                     \\
Muslims       & muslim, arab, muhammad,   & muslim, arab, muhammad, \\ & muhammed, islam, hijab,  & muhammed, islam, hijab,     \\
& sharia & sharia \\
Jews          & -jew(i/s)-, heeb- , sikey-,                           & -jew(i/s)-, heeb- , sikey-, \\ & -zionis-, -semit-  & -zionis-, -semit-                         \\
Liberals      & antifa, libtard, communist,      & antifa, libtard, communist, \\ & socialist, leftist, liberal,  & socialist, leftist, liberal,     \\ & democrat & democrat \\
Conservatives & altright, alt-right,  & altright, alt-right, \\ & cuckservative, trumpster, & cuckservative, trumpster,\\ 
 & conservative, republican & conservative, republican \\ \midrule 
\end{tabular}}
\caption{Keywords used in our data filtering process. The use of more emotionally laden terms is justified by their low occurrence compared to more common terms just to ensure a more diverse dataset.}
\label{tab:keywords}
\end{table}

\subsection{Description of the annotation options}
\label{sec:mturkq1}
\begin{figure*}[]
    \centering
    \includegraphics[width=0.88\textwidth]{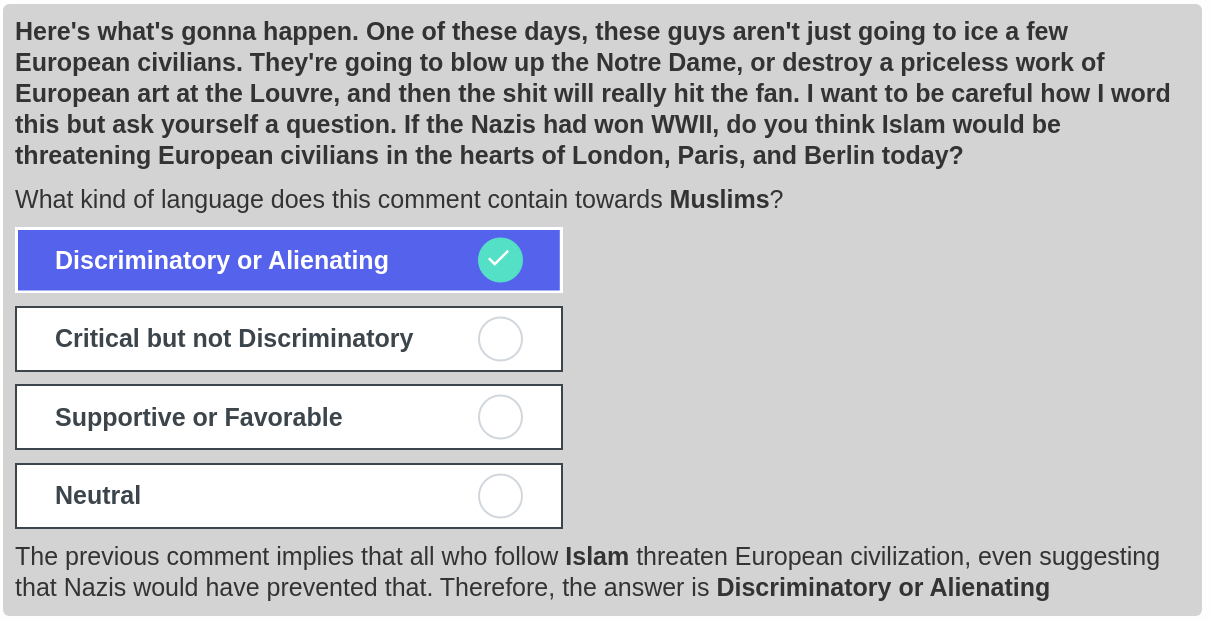}
    \caption{Example of the first question as reference presented to MTurk annotators.}
    \label{fig:questionusvsthem}
\end{figure*}

\paragraph{Discriminatory or Alienating.} Annotators were asked to mark this in case the comment was either, (A) alienating or portraying a social group as negative, (B) a threat, danger or peril to society, (C) trying to ridicule it and attack that group as lesser or worthless. 

\paragraph{Critical but not Discriminatory.} In case the comment was critical, but not to the extent of the first option, annotators were asked to mark this option.

\paragraph{Supportive or Favorable.} This answer refers to comments expressing support towards that group, by defending it or praising it.

\paragraph{Neutral.} This option was offered in case none of the above applied, either because the group was only mentioned but the comment was not addressed at them, or there was no opinion whatsoever expressed towards the group, such as expressing purely factual information.

\begin{figure*}[]
    \centering
    \includegraphics[width=0.88\textwidth]{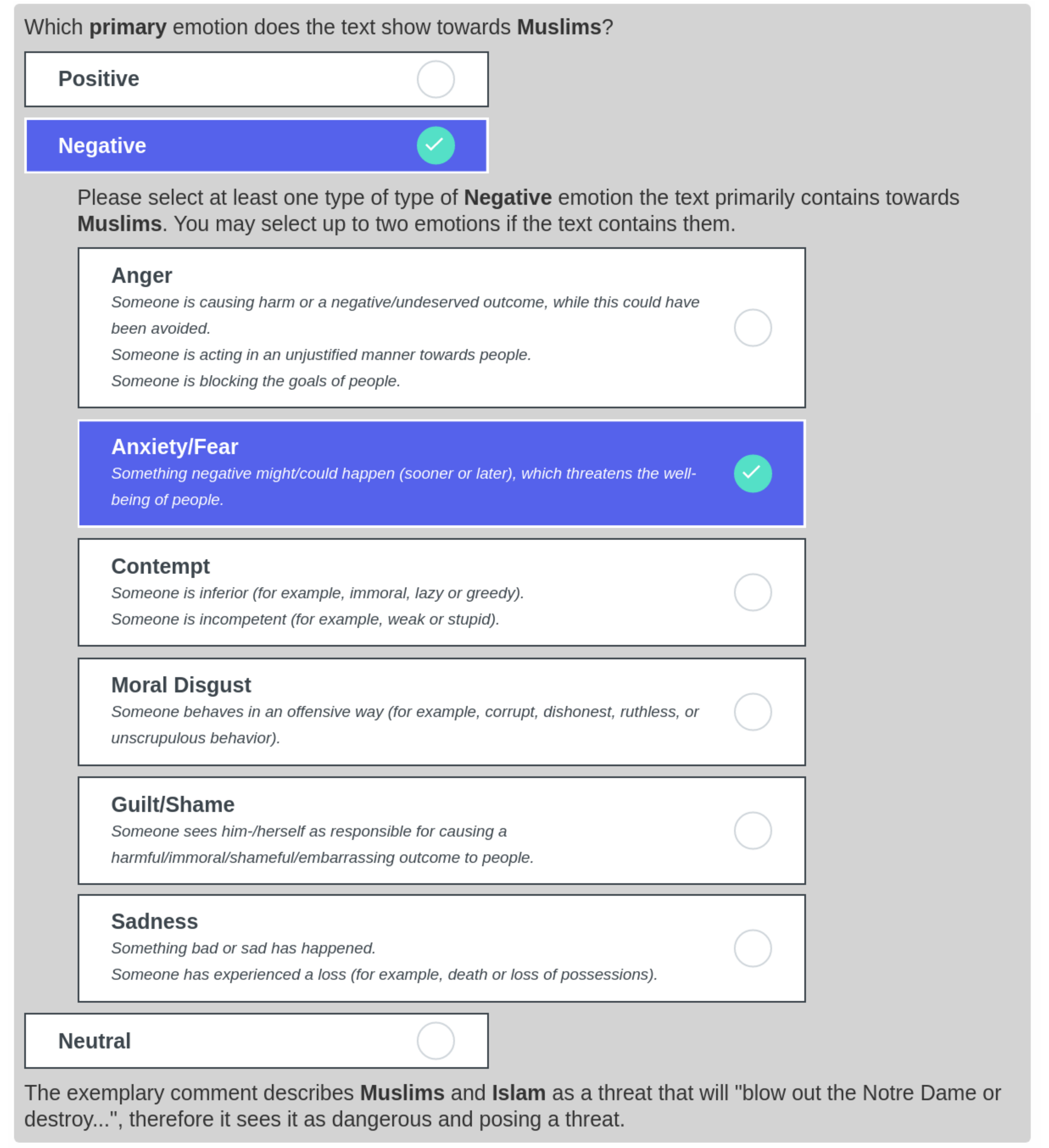}
    \caption{Example of the second question as reference presented to MTurk annotators.}
    \label{fig:questionemotions}
\end{figure*}

Annotators were first asked to select whether the comment showed a `Positive', `Negative' or `Neutral' sentiment towards the specified group. With this approach, we intended to simplify the task and guide annotators, which then were offered to choose from 6 positive or 6 negative emotions according to sentiment they initially chose. In case annotators selected Neutral no further options were provided. The descriptions for each emotion were: 

\paragraph{Positive emotions:}
\noindent
\textbf{Gratitude} \textit{Someone is doing/causing something good or lovely.}\\
\noindent
\textbf{Happiness/Joy}\footnote{Referred to as Happiness for simplicity} \textit{Something good is happening.\\
Something amusing or funny is happening.}\\
\noindent
\textbf{Hope} \textit{Something good/better might happen (sooner or later).}\\
\noindent
\textbf{Pride} \textit{Someone is taking credit for a good achievement.} \\
\noindent
\textbf{Relief} \textit{Something bad has changed for the better.}\\
\noindent
\textbf{Sympathy} \textit{Someone shows support or devotion.}
\paragraph{Negative emotions:}
\textbf{Anger.}  \textit{Someone is causing harm or a negative/undeserved outcome, while this could have been avoided.}\\
\textit{Someone is acting in an unjustified manner towards people.}\\
\textit{Someone is blocking the goals of people.}\\
\noindent
\textbf{Anxiety/Fear} \textit{Something negative might/could happen (sooner or later), which threatens the well-being of people.} \\
\noindent
\textbf{Contempt} \textit{Someone is inferior (for example, immoral, lazy or greedy).\\
Someone is incompetent (for example, weak or stupid).}\\
\noindent
\textbf{Sadness} \textit{Something bad or sad has happened.\\
Someone has experienced a loss (for example, death or loss of possessions).}\\
\noindent
\textbf{Moral Disgust}\footnote{Referred to as Disgust for simplicity} \textit{Someone behaves in an offensive way (for example, corrupt, dishonest, ruthless, or unscrupulous behavior).} \\
\noindent
\textbf{Guilt/Shame}\footnote{Referred to as Guilt for simplicity} \textit{Someone sees him-/herself as responsible for causing a harmful/ immoral/ shameful/ embarrassing outcome to people.}\\

\subsection{Reliability}
\label{app:reliability}
%\begin{wrapfigure}{r}{0.5\textwidth}
\begin{figure}[h]
    \centering
    \includegraphics[width=\columnwidth]{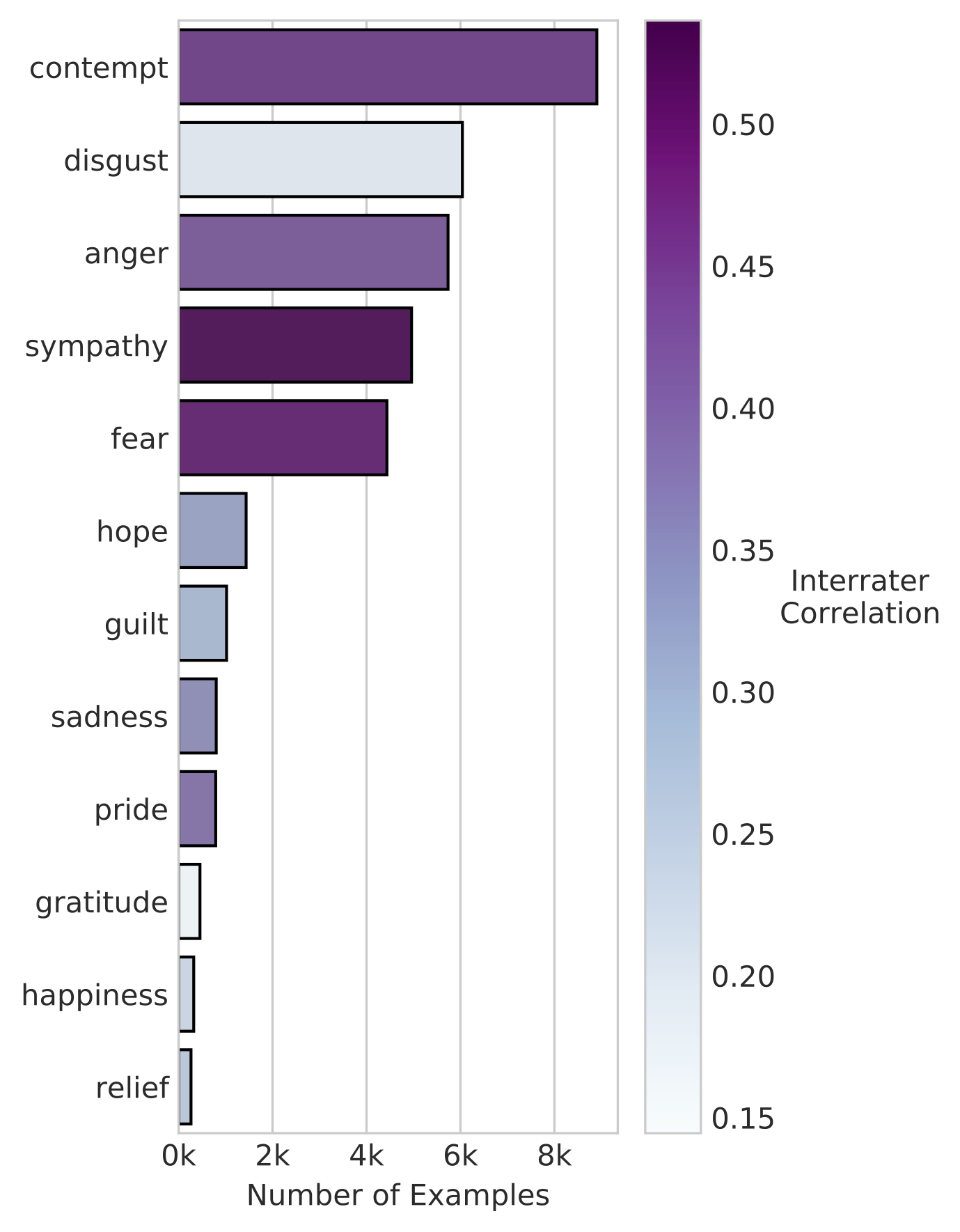}
    \caption{Number of annotations per \\emotions and the inter-rater correlation.}
    \label{fig:correla_dist}
\end{figure}
%\end{wrapfigure}
We also applied the algorithm called \textit{Leave-One-Rater-Out} PPCA \citet{cowen2019}, using Bonferroni correction on p-values. \textit{Principal Preserved Component Analysis} (PPCA) finds principal components which instead of preserving variance within a single dataset as conducted in PCA, preserve the cross-covariance between two different datasets, in our case being a comparison between annotations by one rater and a random set of other raters. In this manner, we can assess the degree of agreement and whether all component dimensions are significant, indicating significant emotion dimensions to be preserved. In our setup, the largest p-value for a dimension was $1.2e-03$, with all other dimensions showing much smaller values. This supports the idea that our emotion dimensions are significant, in order to be kept.
\newpage
\subsection{Data analysis}
\label{sec:appanalysis}

\begin{figure}[h!]
    \centering
    \includegraphics[width=\columnwidth]{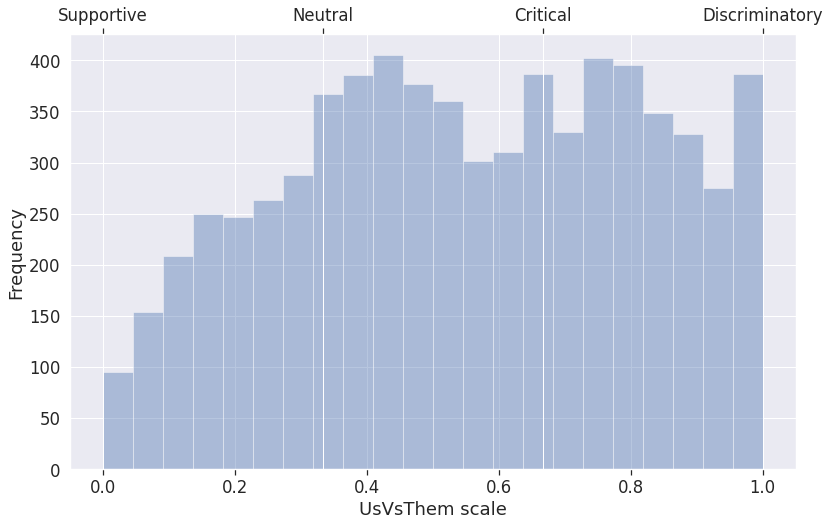}
    \caption{Distribution for the \textit{UsVsThem} scale. Values closer to 0 are more supportive towards the target group, while higher values indicate a higher degree of criticism or eventually discrimination.}
    \label{fig:UsVsThemRegression}
\end{figure}

\begin{table}[h!]
\centering
\resizebox{\columnwidth}{!}{\begin{tabular}{lllllll}\toprule
Predictor          & Sum of & \textit{df} & Mean & \textit{F} & \textit{p} & partial \\
 & Squares &  & Square &  & & $\eta^2 $   \\\midrule
(Intercept)        & 2582.47                            & 1                               & 2582.47                         & $46\times10^3$                     & 0.000                              &          \\
Groups        & 22.05                              & 5                               & 4.41                            & 78.73                          & 0.000                              & 0.04                                  \\
Bias               & 4.82                               & 4                               & 1.21                            & 21.52                          & 0.000                              & 0.01                                 \\
Groups x Bias & 13.82                              & 20                              & 0.69                            & 12.33                          & 0.000                              & 0.03                                 \\
Error              & 492.63                             & 8794                            & 0.06                            &                                &                                &    \\\midrule
\end{tabular}}
\caption{Two-way ANOVA test.}
\label{tab:Anova2way}
\end{table}

\begin{table*}[t!]
\resizebox{2\columnwidth}{!}{\begin{tabular}{lllllll}\toprule
                      & \textbf{Conservatives} & \textbf{Liberals} & \textbf{Immigrants} & \textbf{Refugees} & \textbf{Jews} & \textbf{Muslims} \\\midrule
\textbf{left}         & $0.668^{rc,\,r}$    & $0.540^{rc,\,r}$                & $0.471^{r}$    & $0.524^{r}$              & 0.433           & $0.554^{rc,\,r}$            \\
\textbf{centre-left}  & $0.669^{rc,\,r}$    & $0.513^{rc,\,r}$                & $0.447^{r}$   & $0.507^{r}$              & 0.422           & $0.556^{rc,\,r}$            \\
\textbf{centre}       & $0.646^{rc,\,r}$    & $0.487^{rc,\,r}$                & $0.516^{r}$   & $0.541^{r}$              & 0.452           & $0.573^{rc,\,r}$      \\
\textbf{centre-right} & $0.555^{c,\, lc,\, l}$  & $0.602^{c,\, lc,\, l}$      & $0.497^{r}$    & 0.557             & 0.433           & $0.682^{c,\, lc,\, l}$            \\
\textbf{right}        & $0.543^{c,\, lc,\, l}$       & $0.638^{c,\, lc,\, l}$ & $0.646^{rc,\,c,\,lc,\,l}$               & $0.625^{c,\,lc,\,l}$            & 0.467           & $0.696^{c,\, lc,\, l}$  \\  \midrule       
\end{tabular}}
\caption{Mean \textit{UsVsThem} Regression scale for each group and bias. Statistical significance is shown as super-indexes, in case the mean is statistically different with other biases for that group.  $^{l}$ left, $^{lc}$ centre-left, $^{c}$ centre, $^{rc}$ centre-right, $^{r}$ right. Tested using Tukey HSD test.}
\label{tab:biasRegression}
\end{table*}

\begin{figure*}[t]
    \centering
    \def\svgwidth{2\columnwidth}
    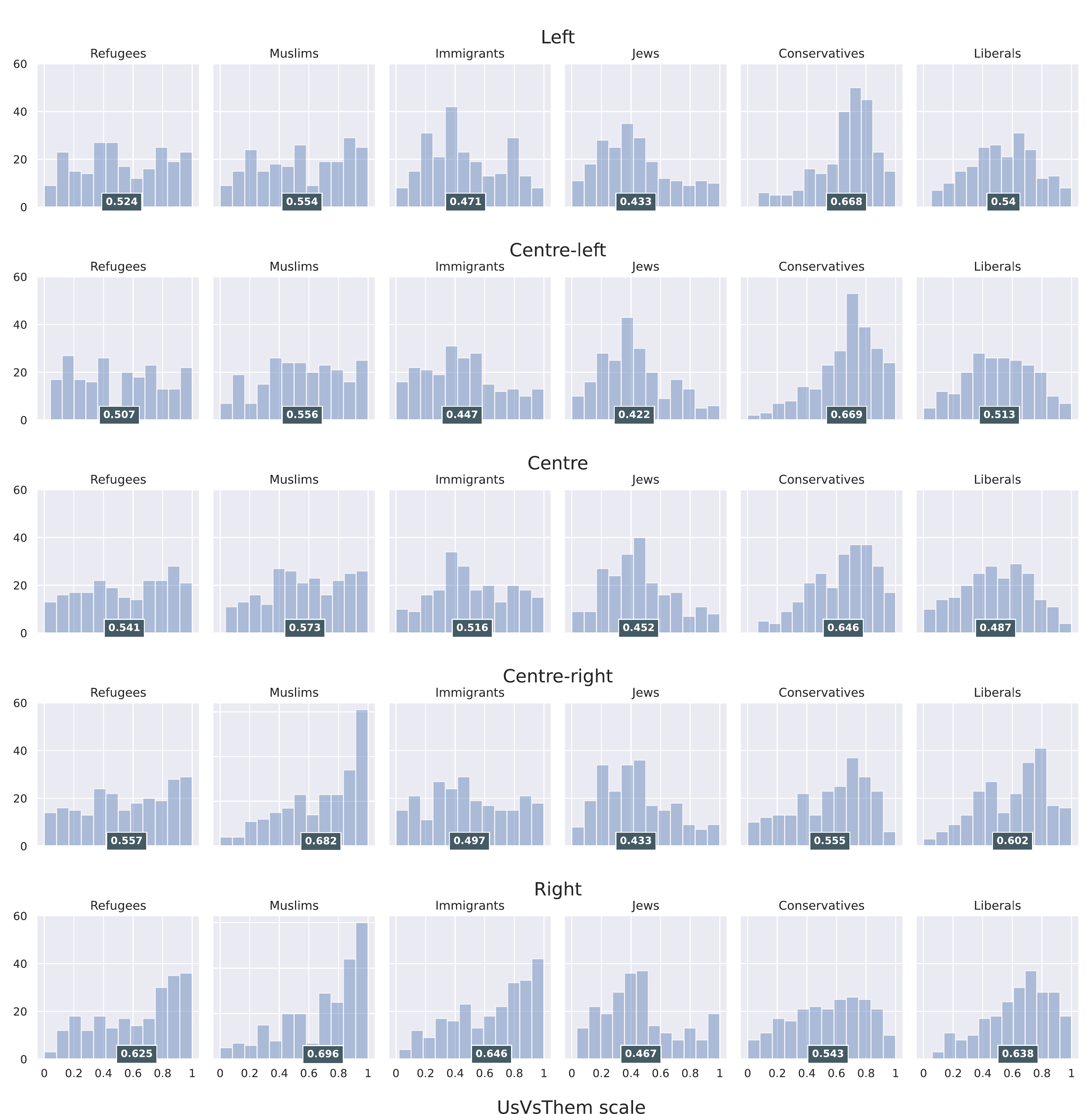
    \caption{Distribution for the \textit{UsVsThem} scale per social group and bias. The mean for the scale is shown at the x axis.}
    \label{fig:UsVsThemRegressionGroupsBiasCombined}
\end{figure*}

\begin{figure*}[ht!]
    \centering
    \includegraphics[width=2\columnwidth]{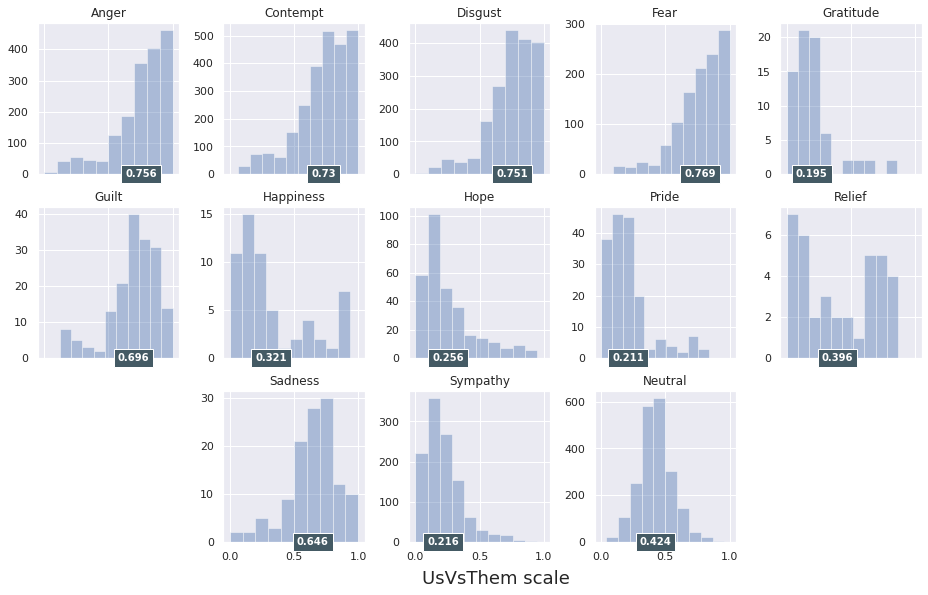}
    \caption{\textit{UsVsThem} scale for each emotion.}
    \label{fig:emotionsUsVsThem}
\end{figure*}

\begin{table*}[h!]
\resizebox{\textwidth}{!}{\begin{tabular}{lrrrrrrrrrrrrr} \toprule
                       & \textbf{Anger} & \textbf{Contempt} & \textbf{Disgust} & \textbf{Fear} & \textbf{Gratitude} & \textbf{Guilt} & \textbf{Happiness} & \textbf{Hope} & \textbf{Pride} & \textbf{Relief} & \textbf{Sadness} & \textbf{Sympathy} & \textbf{Neutral} \\ \midrule
\textbf{Conservatives} & 28.5\% & 47.7\% & 44.4\% & 8.9\%  & 1.3\% & 4.5\% & 1.4\% & 4.2\% & 4.1\% & 0.5\% & 2.4\% & 6.9\%  & 23.7\% \\
\textbf{Liberals}      & 22.4\% & 40.7\% & 28.3\% & 8.4\%  & 1.3\% & 3.1\% & 0.8\% & 7.9\% & 2.9\% & 0.4\% & 1.6\% & 6.4\%  & 32.6\% \\
\textbf{Jews}          & 17.4\% & 22.8\% & 16.9\% & 8.2\%  & 0.5\% & 1.3\% & 0.6\% & 2.3\% & 3.1\% & 0.6\% & 1.5\% & 23.8\% & 44.4\% \\
\textbf{Muslims}       & 31.4\% & 39.2\% & 30.0\% & 26.1\% & 0.8\% & 2.5\% & 0.9\% & 2.0\% & 1.9\% & 0.3\% & 1.5\% & 15.1\% & 26.8\% \\
\textbf{Immigrants}    & 23.5\% & 33.4\% & 18.8\% & 21.5\% & 1.5\% & 1.4\% & 0.8\% & 4.3\% & 2.2\% & 0.6\% & 1.6\% & 23.4\% & 30.9\% \\
\textbf{Refugees}      & 26.9\% & 37.3\% & 21.8\% & 25.8\% & 0.7\% & 2.0\% & 0.6\% & 6.2\% & 1.0\% & 0.8\% & 2.0\% & 24.1\% & 25.3\% \\ \bottomrule       
\end{tabular}}
\caption{Percentages of comments within each social group per emotion.}
\label{tab:groupsemotions}
\end{table*}

\begin{table*}[h!]
\resizebox{\textwidth}{!}{\begin{tabular}{lrrrrrrrrrrrrr} \toprule
                      & \textbf{Anger} & \textbf{Contempt} & \textbf{Disgust} & \textbf{Fear} & \textbf{Gratitude} & \textbf{Guilt} & \textbf{Happiness} & \textbf{Hope} & \textbf{Pride} & \textbf{Relief} & \textbf{Sadness} & \textbf{Sympathy} & \textbf{Neutral} \\ \midrule
\textbf{left}         & 22.5\% & 35.8\% & 25.8\% & 15.1\% & 0.7\% & 2.4\% & 0.6\% & 4.9\% & 2.5\% & 0.5\% & 2.0\% & 18.6\% & 31.5\% \\
\textbf{centre-left}  & 21.1\% & 34.1\% & 27.0\% & 13.5\% & 1.2\% & 2.6\% & 0.7\% & 5.0\% & 2.1\% & 0.4\% & 1.8\% & 18.1\% & 32.6\% \\
\textbf{centre}       & 24.4\% & 35.2\% & 26.1\% & 15.0\% & 1.5\% & 2.1\% & 1.1\% & 5.0\% & 2.7\% & 0.7\% & 1.8\% & 15.1\% & 32.8\% \\
\textbf{centre-right} & 25.9\% & 36.3\% & 26.5\% & 17.0\% & 1.2\% & 2.4\% & 0.9\% & 3.9\% & 3.2\% & 0.7\% & 1.2\% & 17.8\% & 29.4\% \\
\textbf{right}        & 31.5\% & 43.4\% & 29.0\% & 21.9\% & 0.6\% & 2.9\% & 0.9\% & 3.6\% & 2.2\% & 0.4\% & 2.1\% & 13.5\% & 26.4\% \\ \bottomrule
\end{tabular}}
\caption{Percentages of comments within bias in the news source per emotion.}
\label{tab:biasemotions}
\end{table*}
\clearpage

\subsection{Analysis}
\label{sec:appendixresults}
% \vspace{-10em}
%\begin{wrapfigure}{r}{0.5\textwidth}
\begin{figure}[h!]
    \centering
    \includegraphics[width=\columnwidth]{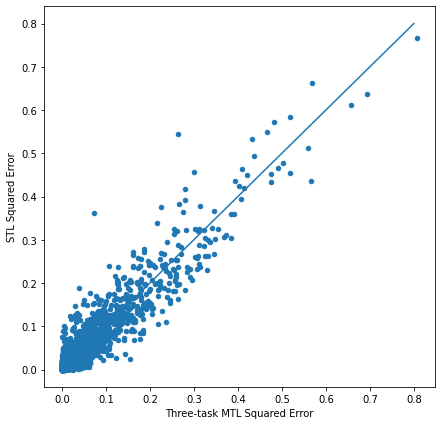}
    \caption{Squared error for STL and three-task MTL}
    \label{fig:errors_stl_mtl}
\end{figure}

% \vspace{-15em}

\begin{table*}[!h] \centering
\resizebox{1\textwidth}{!}{\begin{tabular}{llllllll}
\toprule
$n$ &\multicolumn{7}{l}{\textbf{Reddit Comment}} \\ 
\midrule
1 & \multicolumn{7}{p{\textwidth}}{I can't believe this bullshit. It's literally come down to picking between letting refugees sleep in your bed and fuck your wife and daughter or you're a racist hate monger. Literally no point on the spectrum exists between the two ends. } \\
\midrule
 &\textbf{Label}& \textbf{MTL, E. \& G.} & \textbf{MTL, E.} & \textbf{MTL, G.} &	\textbf{STL} & \textbf{Group}& \textbf{Emotions}\\ 
\midrule
& $0.920$ & $0.646$ & $0.752$ & $0.376$ & $0.655$ & Refugees & Anger \& Fear \\
\midrule
\midrule
2 & \multicolumn{7}{p{\textwidth}}{As a legal immigrant, the newfound term `undocumented immigrant' annoys the heck out of me. They're illegal aliens. Stop trying to sugarcoat it. It took me years to move here legally, and I resent those who chose to do it illegally. The process is long but it is fair. Come in through the front door, not the backdoor.} \\
\midrule
 &\textbf{Label}& \textbf{MTL, E. \& G.} & \textbf{MTL, E.} & \textbf{MTL, G.} &	\textbf{STL} & \textbf{Group}& \textbf{Emotions}\\ 
\midrule
& $0.746$ & $0.661$ & $0.530$ & $0.577$ & $0.436$ & Immigrants & Anger \& Disgust\\
\midrule
\midrule
3 & \multicolumn{7}{p{\textwidth}}{By every moral or ethical standard, it is your duty to refuse orders to “defend” the US from these migrants. History will look kindly upon you if you do. There are thousands, if not millions, of us who will support your decision to lay your weapons down.}\\
\midrule
 &\textbf{Label}& \textbf{MTL, E. \& G.} & \textbf{MTL, E.} & \textbf{MTL, G.} &	\textbf{STL} & \textbf{Group}& \textbf{Emotions}\\ 
\midrule
& $0.17$ & $0.923$ & $0.856$ & $0.884$ & $0.83$ & Immigrants & Sympathy \& Hope \\
\midrule
\midrule
4 & \multicolumn{7}{p{1.005\textwidth}}{I was about to be shocked, until i thought about the god damn state of the world, the western world is at the moment at almost the same state, where at least a large minority wish the same thing of the Muslims. That and god damn people THERE IS MILLIONS OF MUSLIMS NOT EVERYONE THINKS THIS WAY!} \\
\midrule
 &\textbf{Label}& \textbf{MTL, E. \& G.} & \textbf{MTL, E.} & \textbf{MTL, G.} &	\textbf{STL} & \textbf{Group}& \textbf{Emotions}\\ 
\midrule
& $0.099$ & $0.847$ & $0.833$ & $0.882$ & $0.815$ & Muslims & Sympathy\\
\midrule
\midrule
5 & \multicolumn{7}{p{\textwidth}}{The Democrats are the ones preventing people? That's funny. Who are the lawmakers in the state legislatures that are constantly scheming up roundabout ways to defund planned parenthood and completely outlaw abortion access, despite a large majority of Americans supporting at least some degree of abortion? Hint: they're not Dems.}  \\
\midrule
 &\textbf{Label}& \textbf{MTL, E. \& G.} & \textbf{MTL, Emo.} & \textbf{MTL, Groups} &	\textbf{STL} & \textbf{Group} & \textbf{Emotions}\\ 
\midrule
& $0.059$ & $0.78$ & $0.75$ & $0.766$ & $0.734$ & Liberals & Sympathy \\
\bottomrule
\end{tabular}}
\caption{1 and 2 are examples of predictions for comments with high values on the \textit{UsVsThem} scale where MTL models showed an improvement over STL, 3 and 4 are examples of ambiguous and challenging comments and 5 is an example with mentions to more than one group with high error predictions. Predictions are averages of all 10 seeds predictions for each model.}
\label{tab:examplesTotal}
\end{table*}

\begin{figure*}[t!]
    \centering
    \includegraphics[width=1.8\columnwidth]{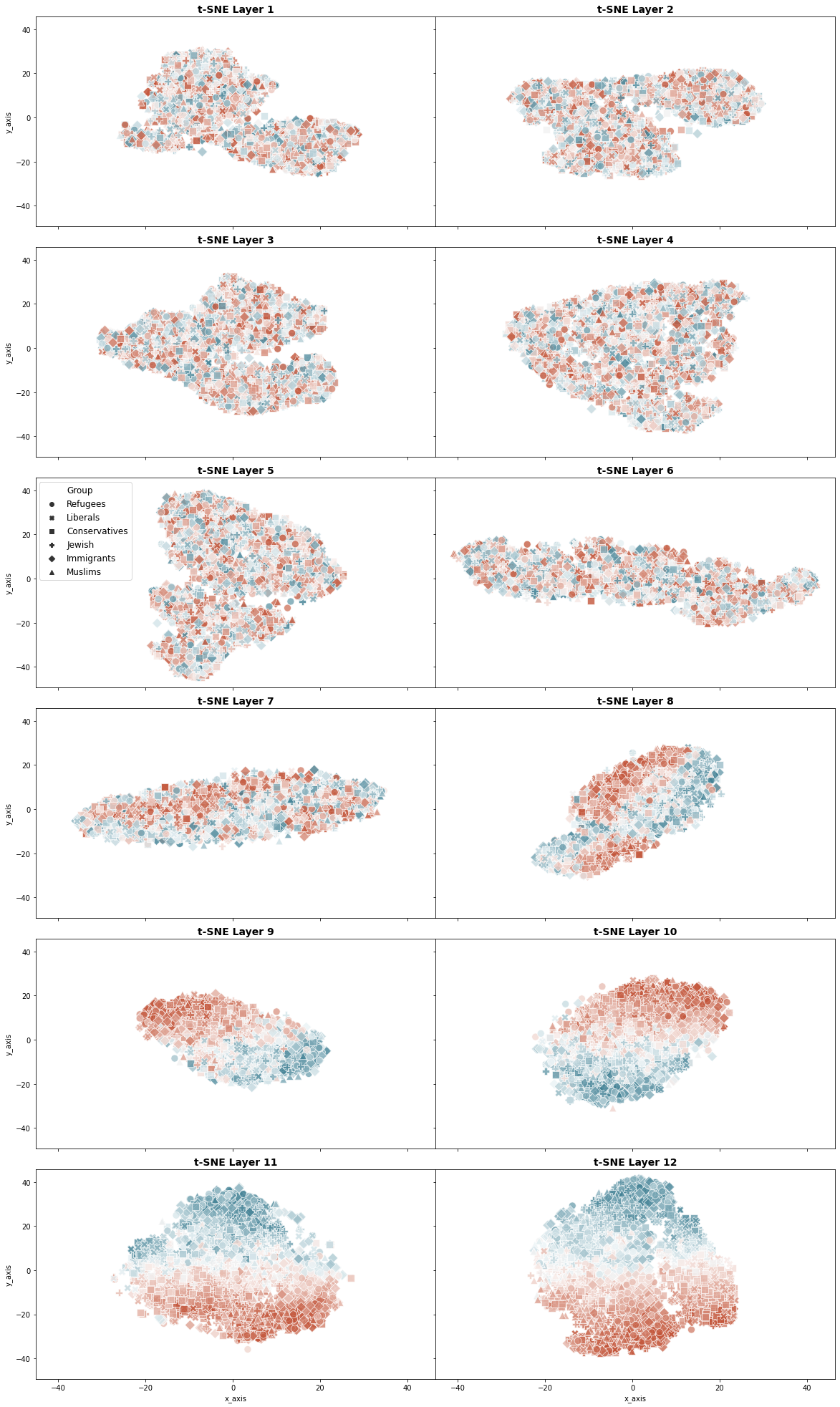}
    \caption{Hidden representations at each layer of the Transformer model for the single task model. Red represents a value closer to 1 in the \textit{UsVsThem} scale and blue closer to 0.}
    \label{fig:layers_stl}
\end{figure*}
\begin{figure*}[t!]
    \centering
    \includegraphics[width=1.78\columnwidth]{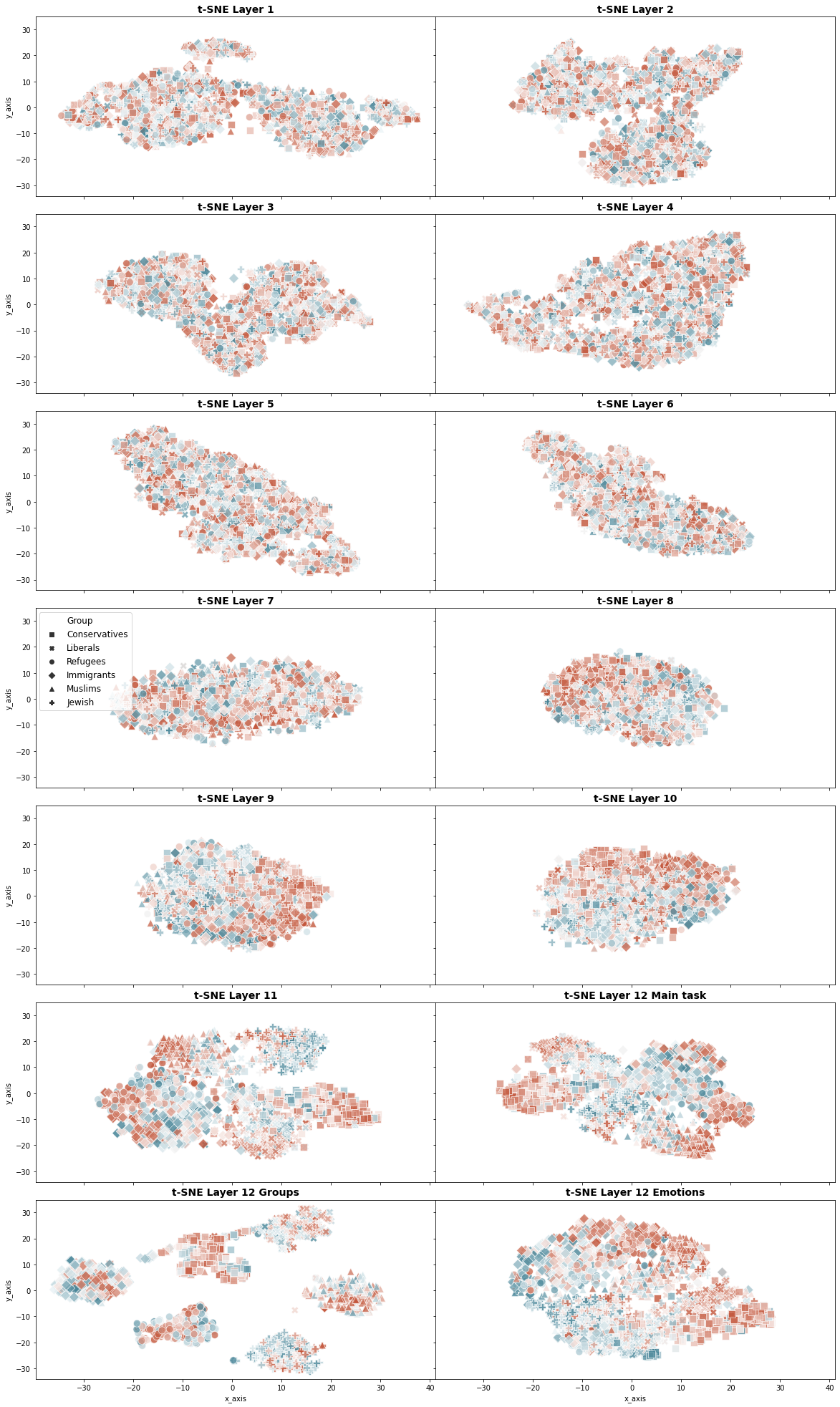}
    \caption{Hidden representations at each layer of the Transformer model for the three-task MTL. The last plots show the task specific Transformer layer output. Red represents a value closer to 1 in the \textit{UsVsThem} scale and blue closer to 0.}
    \label{fig:layers_three}
\end{figure*}
\clearpage
\begin{sidewaysfigure*}[ht!]
    \centering
    \includegraphics[width=\columnwidth]{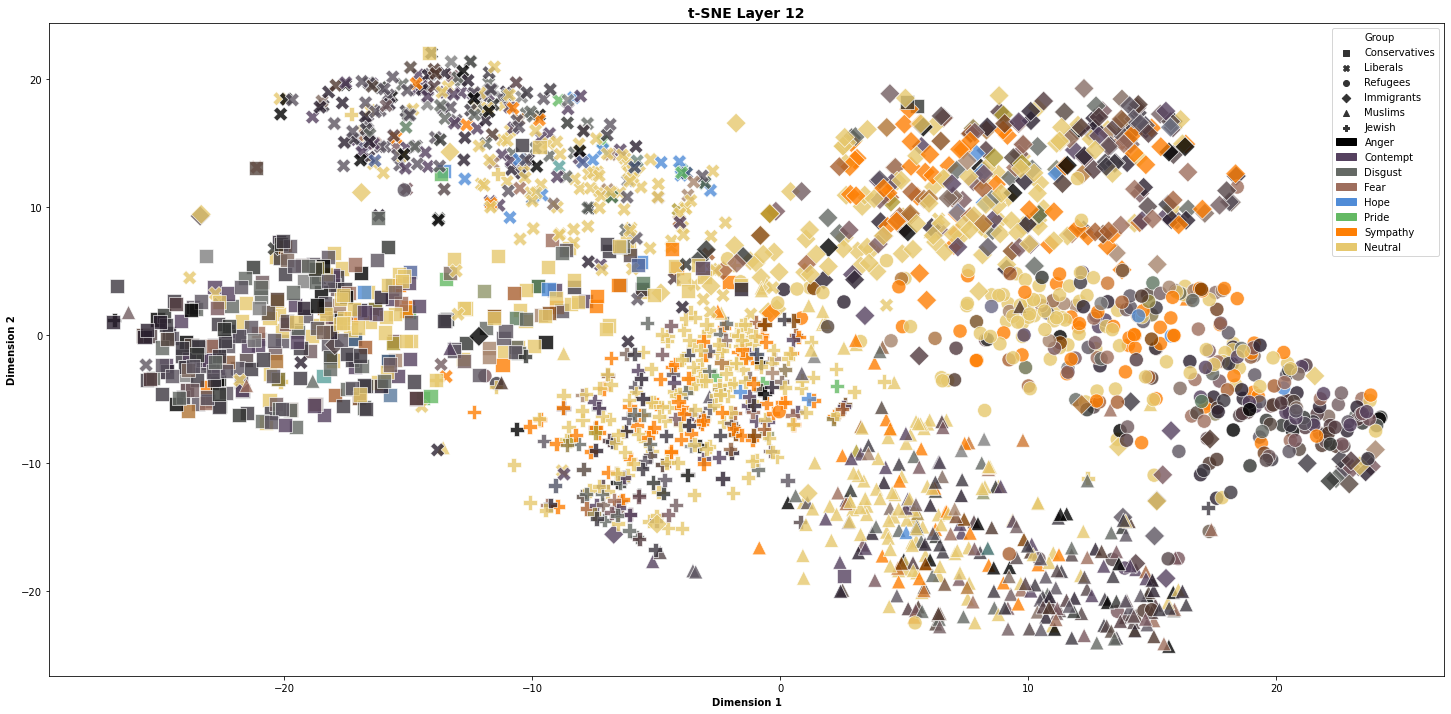}
    \caption{Hidden representations for the three-task MTL main task specific layer. Emotions are represented by colours. Comments with more than one emotion show an average of the colours.}
    \label{fig:final_main_emo}
\end{sidewaysfigure*}
\clearpage
\begin{sidewaysfigure*}[ht!]
    \centering
    \includegraphics[width=\columnwidth]{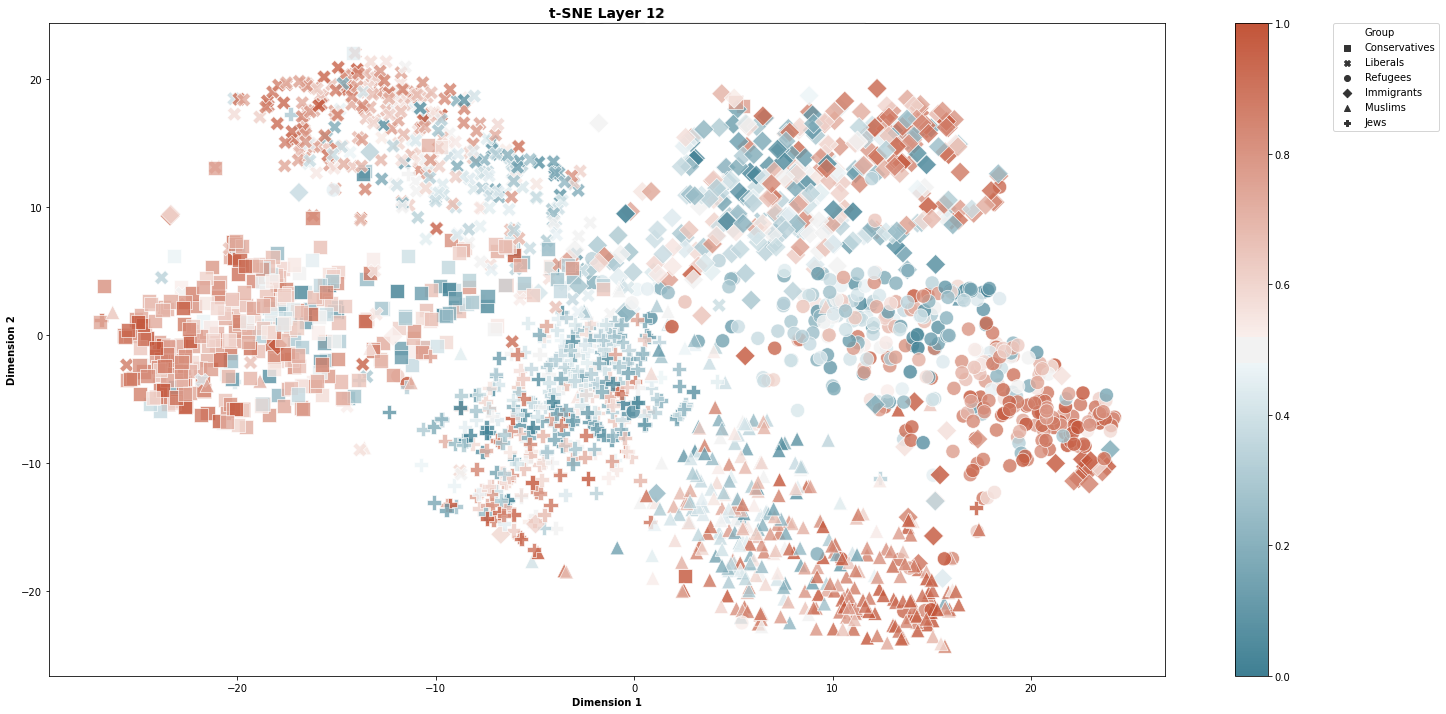}
    \caption{Hidden representations for the three-task MTL main task specific layer. Red represents a value closer to 1 in the \textit{UsVsThem} scale and blue closer to 0.}
    \label{fig:final_layer_main}
\end{sidewaysfigure*}
\clearpage
\begin{sidewaysfigure*}[t!]
    \centering
    \includegraphics[width=\columnwidth]{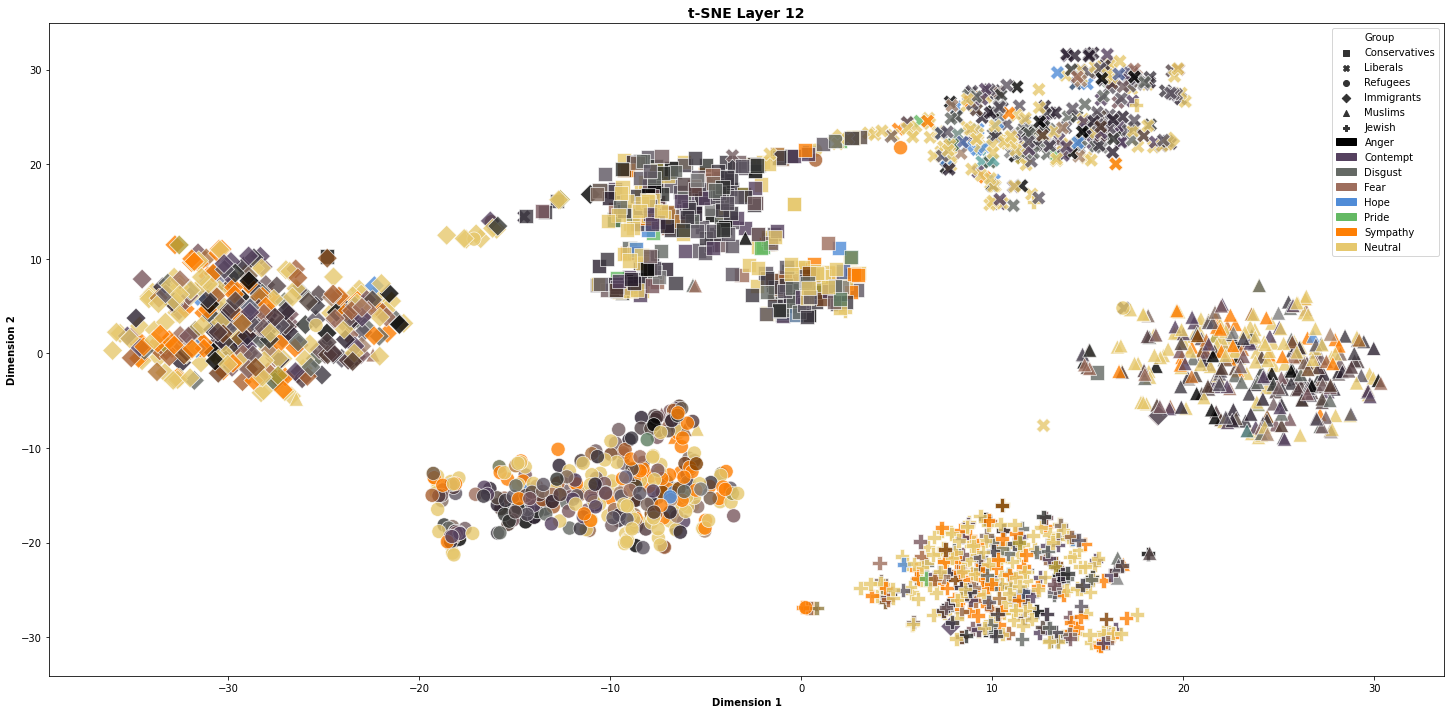}
    \caption{Hidden representations for the three-way MTL group identification specific layer. Emotions are represented by colours. Comments with more than one emotion show an average of the colours.}
    \label{fig:final_group_emo}
\end{sidewaysfigure*}
\clearpage
\begin{sidewaysfigure*}[t!]
    \centering
    \includegraphics[width=\columnwidth]{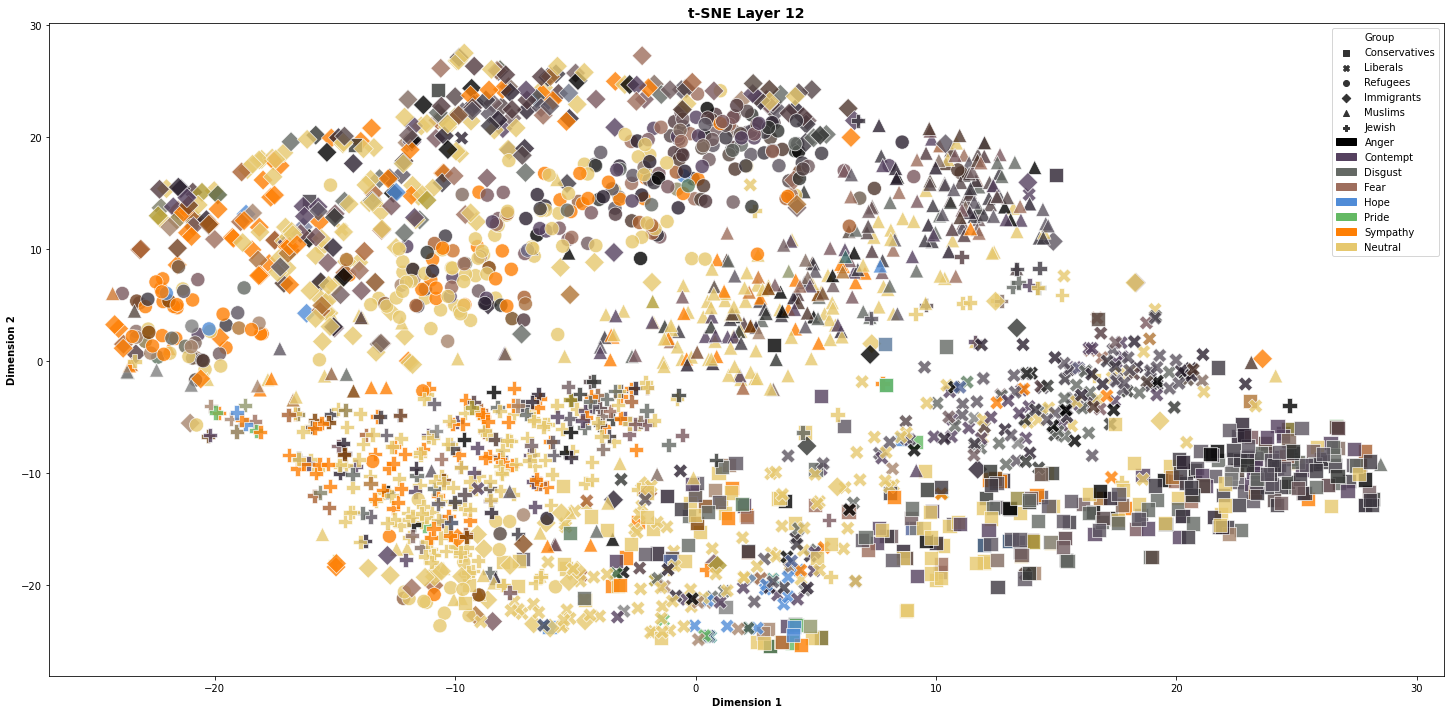}
    \caption{Hidden representations for the three-way MTL emotion specific layer. Emotions are represented by colours. Comments with more than one emotion show an average of the colours.}
    \label{fig:layers_emotions}
\end{sidewaysfigure*}
\end{document}